\documentclass[english,final]{siamonline220329}
\usepackage[T1]{fontenc}
\usepackage[latin9]{inputenc}
\usepackage{verbatim}
\usepackage{float}
\usepackage{textcomp}
\usepackage{amsmath}
\usepackage{graphicx}

\makeatletter

\providecommand{\tabularnewline}{\\}
\floatstyle{ruled}
\newfloat{algorithm}{tbp}{loa}
\providecommand{\algorithmname}{Algorithm}
\floatname{algorithm}{\protect\algorithmname}

\usepackage{cleveref}
\let\ref\cref
\let\eqref\cref
\let\autoref\cref

\makeatother

\usepackage{babel}
\begin{document}
\title{Image warp preserving content intensity}
\author{Enrico Segre\thanks{Physics Core Facilities, Weizmann Institute of Science, Rehovot, Israel, (\email{enrico.segre@weizmann.ac.il})} }

\maketitle
\headers{Image warp preserving intensity}{Enrico Segre}
\begin{abstract}
An accurate method for warping images is presented. Differently from
most commonly used techniques, this method guarantees the conservation
of the intensity of the transformed image, evaluated as the sum of
its pixel values over the whole image or over corresponding transformed
subregions of it. Such property is mandatory for quantitative analysis,
as, for instance, when deformed images are used to assess radiances,
to measure optical fluxes from light sources, or to characterize material
optical densities. The proposed method enforces area resampling by
decomposing each rectangular pixel in two triangles, and projecting
the pixel intensity onto half pixels of the transformed image, with
weights proportional to the area of overlap of the triangular half-pixels.
The result is quantitatively exact, as long as the original pixel
value is assumed to represent a constant image density within the
pixel area, and as long as the coordinate transformation is diffeomorphic.
Implementation details and possible variations of the method are discussed.
\end{abstract}


\section{Introduction}

Many scientific procedures which make quantitative use of the image
content, in fields which range from microscopy to astronomy, require
images to be transformed and remapped onto deformed coordinates system.
Typical applications include the correction of geometrical aberrations
produced by imaging systems, the mutual registration of scenes recorded
with different optical systems or from different points of view \cite{ZITOVA2003977,Modersitzki2004};
stitching together different images with partial overlap among themselves
\cite{GHOSH20161}, also referred to as multi-frame joint image registration;
the fusion and proper coadding of different images of the same source
fields \cite{Zackay2017a}. In medical imaging and computational anatomy
in particular, to name another application, cross image registration
is always required to properly compare features of compliant soft
tissue. Furthermore, dynamic mapping of image sequences over deforming
templates was used with expressive intent in yesteryears, in a procedure
called ``morphing'' \cite{Wolberg1990,radke_2012} where the appearance
of one object was transformed smoothly into that of another (e.g.
a human figure into an animal) by means of gradual deformation and
blending.

For our purposes, we consider two dimensional images, generically
represented as two dimensional arrays of values of the intensity over
Cartesian grids. The methods for determining the appropriate geometrical
transformation between the source and the target coordinates are varied
and sophisticated \cite{Glasbey1998,Modersitzki2009}, depend on the
task, and are not themselves of concern of this paper. Such methods
may make use of functional relations between the coordinate systems
known a priori, or may rely on the identification of common landmark
features appearing in the images \cite{Beier1992,Lee1998}, either
known from supervised annotation, from model fit or from trained deep
learning (e.g.~\cite{Yang2017,Zhang2020}). In computational anatomy,
for instance, diffeomorphic flow is assumed between source and target
images, and LDDMM \cite{Beg2005} in a number of variants is very
popular. A large body of literature exists on these methods, which
do not need to be reviewed here. Once the functional transformation
which maps the two systems of coordinates is established, a ``best''
way of transforming also the image values is sought. The acception
of ``best'' is sometimes subjective and in many cases depends on
the application: it may refer to a cosmetically pleasing result, to
an optimal way of representing and preserving sharp level transitions
in the destination image, or to the suppression of moiré or aliasing
artifacts. Optimal ways of prefiltering and resampling of transformed
images by means of interpolation are well described in literature
\cite{Parker1983,Amanatiadis2009,Getreuer2011} and implemented in
widely adopted software libraries (e.g.~OpenCV \cite{opencv_library},
scikit \cite{scikit-image}, ImageMagick \cite{imagemagick}) as well
as in graphic applications. In other cases, like in superresolution
imaging and reconstruction \cite{Milanfar2011}, the recovery of realistic,
underresolved image details is achieved relying on a priori subscale
models, or optimal use of information resulting from multiple low
resolution images belonging to a sequence. In this paper, in contrast,
we describe a procedure which is purely intended to preserve the brightness
of the image content across the transformation, even when the images
are not Nyquist sampled, without invoking any help from the image
data itself or from a priori knowledge of structure lost by the process
of image formation. In simple terms, we exactly redistribute the whole
intensity content of the source pixels over the target raster. The
procedure is linear, and amounts to the determination of a reweighting
matrix which projects the pixel values from the source to the destination
image, and most importantly, depends only on the coordinate transform
and not on the image data itself. As such, some variants of the procedure
can be devised from the basic scheme, including one which provides
a stable alternative to image interpolation without ad hoc filtering.

Our procedure implements a rigorous area resampling. The concept is
known even from earlier literature, but does not seem to have received
adequate attention, probably because of its higher computational cost
which hinders its applications, and does not seem to have been pursued
in the general case. Early attempts include that of \cite{Fant1986},
which proposes a fast implementation, based on a scanline decomposition.
Scanline approaches treat the deformation of a raster image by carrying
on some of the intensity content from one pixel to its adjacent in
scan order, and are not proven to be exact for transformations beyond
simple shears. A simpler version of Fant's algorithm, applied only
to raster resizing, goes under the name of pixel mixing \cite{Summers2012}
and was probably implemented in the open first by the \emph{pamscale}
function of the netpbm package \cite{Reinelt2003}. The thesis \cite{Chiang1998}
generalizes the problem, introducing the term ``imaging-consistent
integrating resampler'', taking into account also the point spread
function of the imager, and blurring in image formation. The algorithm
proposed there, though, still falls within the category of separable,
scanline approaches, with a single accumulation register providing
intensity remainders carried over from one pixel to the next. Another
double pass, scanline algorithm is that of \cite{Han2005}. The seminal
thesis \cite{Heckbert1989} discusses the problem, and gives a partial
solution in terms of adaptive local deformation of circular neighborhoods.
This method is also implemented in the popular software package ImageMagick
\cite{Thyssen2012}. Another cognate approach proposed, and employed
specifically for oversampling stacks of dithered astronomical images
while preserving photometry, is ``Drizzle'' \cite{Fruchter2002},
but it relies on empirical factors, and treats both source and destination
pixels as squares. ``Drizzling'' estimates pixel area overlaps using
a sort of a Montecarlo approach, where the randomness is provided
by inter-image pixel shifts. As a procedure, it some offers other
advantages like the possibility of assigning individual quality weights
to each contributing pixel; still it is not general for arbitrary
deformations.

In summary, the existing literature on area resampling concentrated
on the search for ``efficient'' variants of the method, or which
seem to lack generality or exactness, when specifically looking at
the preservation of the photometric intensity. It is our intention
to discuss here a rigorous procedure, and its implementation.

The paper is organized as follows: section \ref{sec:Exact-area-resampling}
describes the geometrical principle of pixel remapping, section \ref{sec:Algorithm-layout}
outlines the algorithm used, section \ref{sec:Warping-examples-and}
demonstrates it, section \ref{sec:Extensions} discusses some variations,
section \ref{sec:Example:-source-photometry} shows the advantage
of area resampling in photometric measurements, and section \ref{sec:Conclusions-and-future}
concludes and outlines future perspectives. Computational details
are included in the appendices: the convention adopted for barycentric
coordinates is given in appendix \ref{app:Barycentric-coordinates};
appendix \ref{app:intersection-area} discusses the problem of intersecting
triangles, and code performance is reported in appendix \ref{app:Computational-performance}.

\section{Exact area resampling by pixel triangulation\label{sec:Exact-area-resampling}}

We start from an intensity image, given as a set of $N_{1}\times M_{1}$
pixel values $I_{1}(i,j)$, representing the cumulative value of some
quantity (for instance, the number of photons impinging the area of
an individual photosensitive element), integrated over the rectangular
pixel $p_{ij}$, defined as the rectangle $x_{i}\le x<x_{i+1}$, $y_{j}\le y<y_{j+1}$,
for $1\le i\le N_{1}$ and $1\le j\le M_{1}$. For simplicity we will
treat here an equispaced coordinate grid, $x_{i}=x_{1}+(i-1)\cdot\Delta x$
and $y_{j}=y_{1}+(j-1)\cdot\Delta y$, though the procedure can be
easily generalized to non-equispaced plaid grids. We consider an a
priori given bijective and differentiable coordinate transformation
$(X,Y)=f(x,y)$. We assume that the underlying intensity density $i_{1}(x,y)$
inside the pixel $p_{ij}$ is uniform, and that $\int_{p_{ij}}i_{1}(x,y)\,dx\,dy=I_{1}(i,j)$.
Therefore, $i_{1}(x,y)=I_{1}(i,j)/\mathcal{A}\left[p_{ij}\right]$,
where $\mathcal{A}\left[p_{ij}\right]=\Delta x\Delta y$ is the area
of the pixel.

\begin{figure}
\begin{centering}
\includegraphics[width=0.65\textwidth]{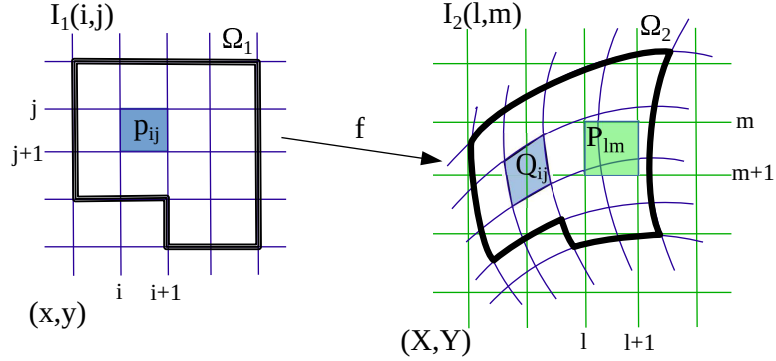}
\par\end{centering}
\caption{\label{fig:Pixel-transform}Pixel transform $p_{ij}\to Q_{ij}$ from
the source to the destination image space, and mapping of closed contours
$\Omega_{1}\to\Omega_{2}$ including specific pixel groups.}
\end{figure}

Our goal is to produce a new image of $N_{2}\times M_{2}$ pixels,
transforming the set of values $I_{1}$ into a new set $I_{2}(l,m)$
on a new equispaced grid $\left\{ \left(X_{l},Y_{m}\right)\right\} $,
with $1\le l\le N_{2}$, $1\le m\le M_{2}$ and spacing $\Delta X$,
$\Delta Y$, in such a way that the cumulative intensity within any
closed contour is preserved by the transformation:
\begin{equation}
\int_{\Omega_{1}}i_{1}(x,y)\,dx\,dy=\int_{\Omega_{2}}i_{2}(X,Y)\,dX\,dY\label{eq:integral-intensity-conservation}
\end{equation}
 for any region $\Omega_{2}=f\left(\Omega_{1}\right)$, and assuming
an underlying transformed intensity density $i_{2}$ in the destination
image. A natural way of achieving this property is to consider the
quadrilateral $Q_{ij}=\left\{ f\left(x_{i},y_{j}\right),f\left(x_{i+1},y_{j}\right),f\left(x_{i+1},y_{j+1}\right),f\left(x_{i},y_{j+1}\right)\right\} $,
which approximates (to second order in $\Delta x$, $\Delta y$) the
transform of the rectangular pixel $p_{ij}$, identified by the set
of its four vertices $\left\{ \left(x_{i},y_{j}\right),\left(x_{i+1},y_{j}\right),\left(x_{i+1},y_{j+1}\right),\left(x_{i},y_{j+1}\right)\right\} $
(see Fig.~\ref{fig:Pixel-transform}). Save for singular or extreme
deformations and coarse griddings which are of little practical interest,
we can tacitly assume that $Q_{ij}$ remains a convex quadrilateral
(concavity would imply a change of sign of the Jacobian of the transformation,
violating the assumption of diffeomorphic transformation). For shorthand,
we write $Q_{ij}\simeq f\left(p_{ij}\right)$, applying $f()$ to
polygons and contours as well as to individual points. Locally, this
scalar density would be transformed as
\begin{equation}
i_{2}(X,Y)=J_{f}\,i_{1}(x,y)=\left|\begin{array}{cc}
\frac{\partial X}{\partial x} & \frac{\partial X}{\partial y}\\
\frac{\partial Y}{\partial x} & \frac{\partial Y}{\partial y}
\end{array}\right|i_{1}(x,y)\,,\label{eq:Jacobian}
\end{equation}
so that, to second order, (\ref{eq:integral-intensity-conservation})
is satisfied for $Q_{ij}$ and hence for any region composed of sets
of pixels of image 1. Within the same approximation, we assume that
the Jacobian $J_{f}$ is constant within $p_{ij}$ and the density
$i_{2}$ constant within $Q_{ij}$. This position allows us to reduce
the change of integration variable in Eq. (\ref{eq:integral-intensity-conservation})
into a problem of decomposition of pixel areas: each fraction of $Q_{ij}$
projected on the destination grid, will contribute to the target intensity
proportionally to its fractional area only.

The idea of considering the shape change of the pixel in the transformation
is not new (see for example of the procedure described in §15.5 of
\cite{Velho2009}); however, our development is different in that
we do not invoke arbitrary interpolations for the reconstruction of
the destination image. The intensity of the rectangular pixel 
\[
P_{lm}=\left\{ \left(X_{l},Y_{m}\right),\left(X_{l+1},Y_{m}\right),\left(X_{l+1},Y_{m+1}\right),\left(X_{l},Y_{m+1}\right)\right\} 
\]
 on the target image is expressed as a sum of contributions
\begin{equation}
I_{2}(l,m)=\sum_{\text{overlaps}}I_{2}^{ij}(l,m)=\sum_{\text{overlaps}}\frac{\mathcal{A}\left[P_{lm}\cap Q_{ij}\right]}{\mathcal{A}\left[Q_{ij}\right]}\,I_{1}(i,j)\,,\label{eq:pixel_overlap_contributions}
\end{equation}
from each of the transformed pixels $Q_{ij}$ of image 1 partially
overlapping with $P_{lm}$ in image 2. The subset of indices $i,j$
to be taken into account is indicated here generically as ``$\text{overlaps}$'';
a criterion for selecting them will be formulated in the following.
In (\ref{eq:pixel_overlap_contributions}), $\mathcal{A}\left[\right]$
indicates the area of the resulting polygon. The procedure involves
therefore two steps: 1) for any given destination pixel $P_{lm}$
identify the set of original pixels $p_{ij}$ whose transform $Q_{ij}$
overlaps with it, and 2) determine the polygonal shape of each intersection
and compute its area.

\begin{figure}
\begin{centering}
\includegraphics[width=0.45\textwidth]{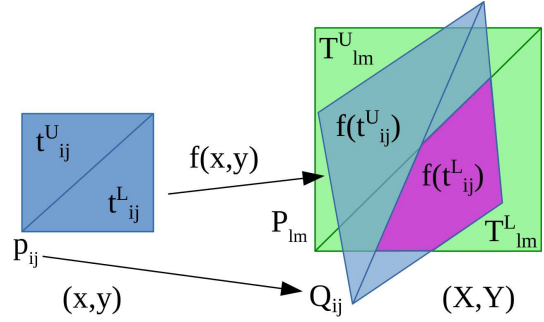}
\par\end{centering}
\caption{\label{fig:Pixel-Decomposition}Decomposition of a source pixel $p_{ij}=t_{ij}^{U}\cup t_{ij}^{L}$
in two triangles in the origin and in the destination space, $Q_{ij}=f\left(p_{ij}\right)=f\left(t_{ij}^{U}\right)\cup f\left(t_{ij}^{L}\right)$,
and its overlap with the destination pixel $P_{lm}=T_{lm}^{U}\cup T_{lm}^{L}$.
The intersection $T_{lm}^{L}\cap f\left(t_{ij}^{L}\right)$, in this
case a pentagon, is highlighted in purple for illustration. }

\end{figure}

The intersection of two convex quadrangles can be, in general, a polygon
with anything between three and eight sides. Algorithms for the intersection
of generic polygons exist in reputable computer geometry packages
(e.g. in CGAL \cite{cgal:fwzh-rbso2-20b}), but their generality comes
as a hindrance for our specialized case, requiring peculiar organized
data structures, and is not necessarily optimal for a fast calculation.
We prefer to simplify the task one step further. We divide both the
origin pixel $p_{ij}$ and the destination pixel $P_{lm}$ in two
triangles, splitting the quadrangles arbitrarily along one of their
two diagonals, for instance $t_{ij}^{U}=\left\{ \left(x_{i},y_{j}\right),\left(x_{i+1},y_{j+1}\right),\left(x_{i},y_{j+1}\right)\right\} $
and $t_{ij}^{L}=\left\{ \left(x_{i},y_{j}\right),\left(x_{i+1},y_{j}\right),\left(x_{i+1},y_{j+1}\right)\right\} $,
and analogously $T_{lm}^{U}$ and $T_{lm}^{L}$ (Fig.~\ref{fig:Pixel-Decomposition}).

The task of identifying intersections between $N_{1}\times M_{1}$
origin and $N_{2}\times M_{2}$ destination pixels, required by Eq.~(\ref{eq:pixel_overlap_contributions}),
is split in that of finding the intersections between four times as
many triangular half pixels. This is still non trivial, as there are
no less than 17 possible, topologically different ways of intersecting
two triangles (Figure \ref{fig:The-seventeen-topologically}), excluding
degenerate cases, as further discussed in appendix \ref{app:intersection-area},
but is definitely a simpler task than for quadrangles.

The same partial intensity $I_{1}(i,j)/2$ is assigned to each of
the two original triangles. Eq.~(\ref{eq:pixel_overlap_contributions})
therefore expands into
\begin{multline}
I_{2}(l,m)=\sum_{\text{overlaps}}\left[\frac{\mathcal{A}\left[T_{lm}^{U}\cap f\left(t_{ij}^{U}\right)\right]+\mathcal{A}\left[T_{lm}^{L}\cap f\left(t_{ij}^{U}\right)\right]}{2\mathcal{A}\left[f\left(t_{ij}^{U}\right)\right]}\right.+\\
\left.\frac{\mathcal{A}\left[T_{lm}^{U}\cap f\left(t_{ij}^{L}\right)\right]+\mathcal{A}\left[T_{lm}^{L}\cap f\left(t_{ij}^{L}\right)\right]}{2\mathcal{A}\left[f\left(t_{ij}^{L}\right)\right]}\right]\,I_{1}(i,j)\label{eq:intensity-triangle-contributions}
\end{multline}

Formally, the transformation between pixel intensities from the one
to the other image can be written as
\begin{equation}
I_{2}(l,m)=\sum_{i,j=1}^{N_{1},M_{1}}B_{lm,ij}I_{1}(i,j)\,,\label{eq:matrix-intensity-transformation}
\end{equation}
 where $B_{lm,ij}$ is the incidence matrix detailing which fraction
of $Q_{ij}$ intersects with $P_{lm}$. In typical cases, in which
pixels of the original and of the destination rasters are of comparable
sizes, this matrix is very sparse.

For transformations in which a source pixel $p_{ij}$ is completely
mapped on pixels on the the destination raster, the property $\sum_{lm}B_{lm,ij}=1$
holds. Conversely, $\sum_{ij}B_{lm,ij}$ gives a discretized representation
of $J_{f}^{-1}$ on the destination raster. 

As an aside, once the transformation from $I_{1}$ to $I_{2}$ has
been computed according to Eq.~(\ref{eq:matrix-intensity-transformation}),
its inverse can be obtained directly inverting the sparse matrix $B$,
for which numerical techniques are well studied. This may be more
advantageous than using the inverse coordinate mapping, if using an
algorithm like the one described in the next section, which exploits
the fact that the source image grid is cartesian.

\section{Algorithm layout\label{sec:Algorithm-layout}}

As resulting from (\ref{eq:matrix-intensity-transformation}), the
intensity transformation between the two rasters reduces to a simple
matrix multiplication, once the matrix elements $B_{lm,ij}$ are computed.
To this extent, the steps sketched in inset \ref{alg:Intensity-transformation}
are required.

\begin{algorithm}[h]
For each pixel of the source image, i.e. iterating on $i$ and $j$:
\begin{enumerate}
\item \label{enu:coord-transform}the coordinates of the vertices of each
original hemipixel $t_{i,j}^{L,U}$, are transformed with $f$
\item \label{enu:overlap-set}the set of hemipixels $\left\{ T_{lm}^{L,U}\right\} $
which have a non empty overlap with $f(t_{ij}^{L,U})$ is determined
\item \label{enu:overlap-area}the area of the intersections between each
of the triangles of this set and each $f(t_{ij}^{L,U})$ in turn,
is found.
\item \label{enu:areas-of-T}the areas $\mathcal{A}\left[f\left(t_{ij}^{U}\right)\right]$
and $\mathcal{A}\left[f\left(t_{ij}^{L}\right)\right]$ are computed.
\item \label{enu:matrix-elements}the relevant contributions are summed
to construct the matrix $B_{lm,ij}$, 
\end{enumerate}
Finally, the image $I_{2}$ is obtained by (\ref{eq:matrix-intensity-transformation}).

\caption{Intensity transformation between $I_{1}$ and $I_{2}$\label{alg:Intensity-transformation}}
\end{algorithm}

Step \ref{enu:coord-transform} is the simple evaluation of a given
function $f$ of the coordinates, and does not need to be described
here. In the terminology of image processing, we are using naturally
here a \emph{forward} mapping between source and destination image.

For step \ref{enu:overlap-set}, all triangles $T$ which have at
least one vertex within the bounding box\\
$\left[\min_{X}\left(Q_{ij}\right),\max_{X}\left(Q_{ij}\right)\right]\otimes\left[\min_{Y}\left(Q_{ij}\right),\max_{Y}\left(Q_{ij}\right)\right]$
are selected ($\otimes$ denoting the Cartesian product of the two
intervals). Since the triangular half pixels $T$ are defined on a
structured grid, they can be indexed in such a way that the criterion
is translated to a simple choice of indices, involving integer arithmetics.
It is algorithmically simpler to use this simplified criterion, which
may sometimes include additional disjoint triangles, than to refine
the search to the subset of triangles which have an actual intersection.
The condition for a positive overlap is not as simple as for instance
the requirement that vertices of $T_{lm}$ fall internally to $f\left(t_{ij}\right)$
or viceversa (figure \ref{fig:The-seventeen-topologically} provides
many counter examples).

The computation of overlap areas is more involved, and performed at
step \ref{enu:overlap-area}. For that, we make due use of barycentric
coordinates \cite{Coxeter1969,EricsonChrister2005RCD} to reference
the position of a point within a given triangle $T$. In barycentric
coordinates, the position of any point $B$ in the plane is determined
by a triple of real numbers $\left(b_{1,}b_{2},b_{3}\right)$. This
system has several properties that come to advantage for topological
tests. With proper normalization, $B$ can be said to be internal
to $T$ if all the three numbers \textbf{$b$ }are positive; $B$
falls on a side of $T$ if one of the three $b$ is null, and coincides
with a vertex of $T$ if two $b$ are simultaneously null. Intersection
points between two segments (in our case, sides of $T$ and of $f(t)$)
are easily computed from the barycentric coordinates of their extremes
(equation \ref{eq:barycrossing}). Since a segment and the side of
a triangle intersect only if the relevant barycentric coordinate of
the extremes have opposite signs, inspection of the signs can also
be used as a flag to avoid unnecessary computation of non existing
crossings. Details are in Appendix \ref{app:Barycentric-coordinates}.

Two possible algorithms for computing the areas of the triangle intersections
needed for step \ref{enu:overlap-area} are described in detail in
Appendix \ref{app:intersection-area}. In our approach we make use
of the one described in \ref{subsec:Brute-force-approach}, which,
albeit possibly slightly less efficient, is of much simpler implementation.

For step \ref{enu:areas-of-T} the area $A$ is elementary obtained
from the vertex coordinates, computing the outer product of two side
vectors, whereas step \ref{enu:matrix-elements} is mechanic.

\section{Warping examples and evaluation\label{sec:Warping-examples-and}}

We provide an example of the area resampling method using an 8 bit
monochrome, $512\times512$ pixels test image (\texttt{boat.512},
from \cite{Weber2018}). The image coordinates are defined so that
$x_{1}=y_{1}=0$ and $x_{512}=y_{512}=1$ ($y$ increasing downwards).
For the sake of illustration we take, as warping transformation,
\begin{equation}
\left(\begin{array}{c}
X\\
Y
\end{array}\right)=f(x,y)=\left(\begin{array}{c}
x+\frac{3\sin\left(2\pi y\right)}{20}\\
y-\frac{3\sin\left(\pi x\right)}{20}
\end{array}\right)\,,\label{eq:wavy}
\end{equation}

which induces no deformation on the sides of the unit square, and
has Jacobian comprised between $0.45<J_{f}<1.65$ . To quantify the
numerical error in the preservation of intensity of the warped image,
we compute the total intensity discrepance
\begin{equation}
\delta=\frac{\sum_{i,j=1}^{N_{1},M_{1}}I_{1}(i,j)-\sum_{l,m=1}^{N_{2},M_{2}}I_{2}(l,m)}{\sum_{i,j=1}^{N_{1},M_{1}}I_{1}(i,j)}\ .\label{eq:delta}
\end{equation}

\begin{figure}
\begin{centering}
\includegraphics[height=6cm]{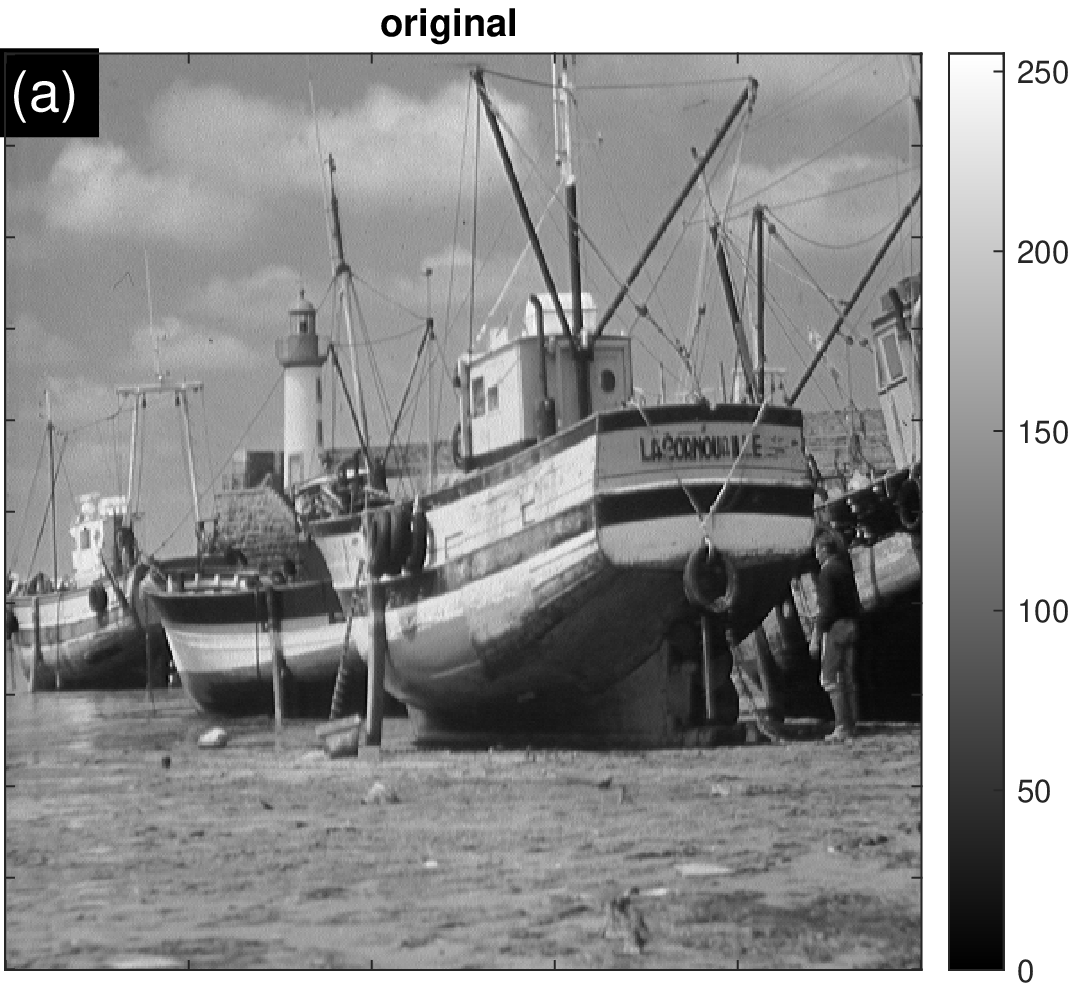}\includegraphics[height=6cm]{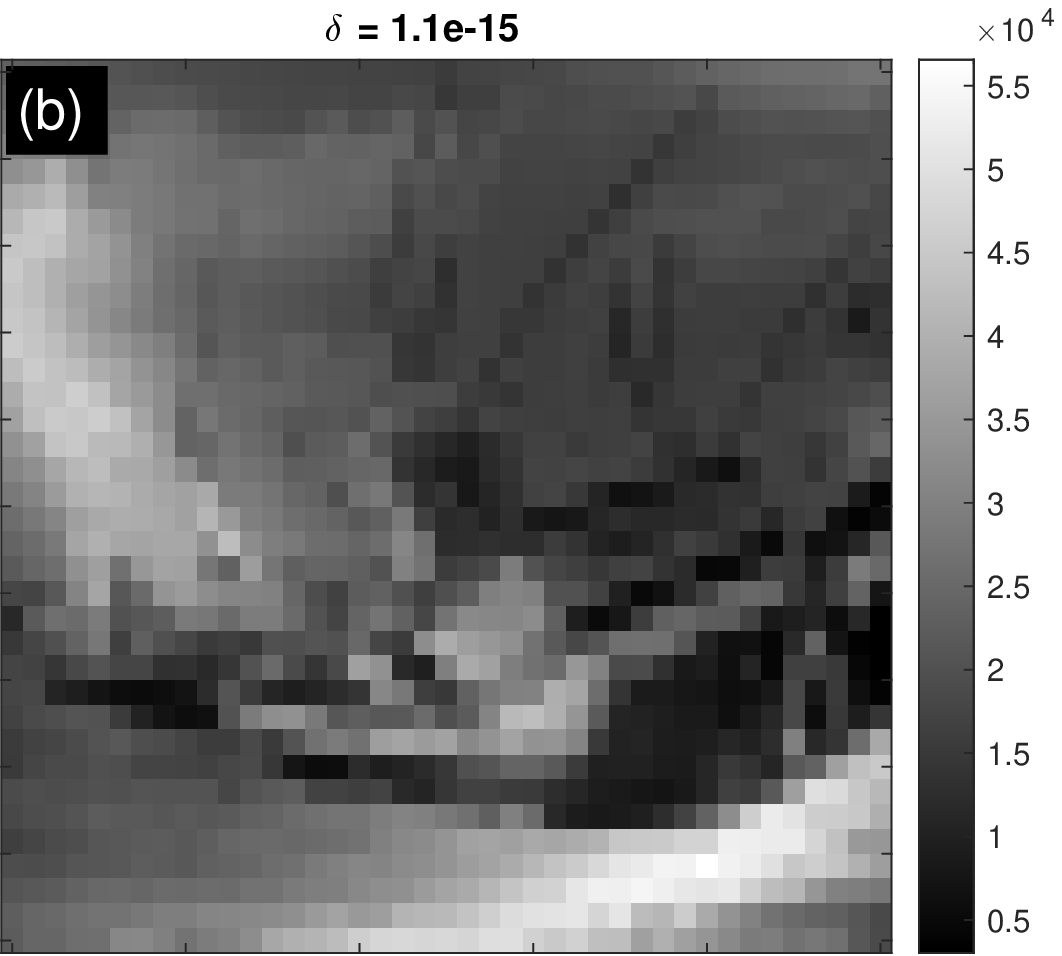}
\par\end{centering}
\begin{centering}
\includegraphics[height=6cm]{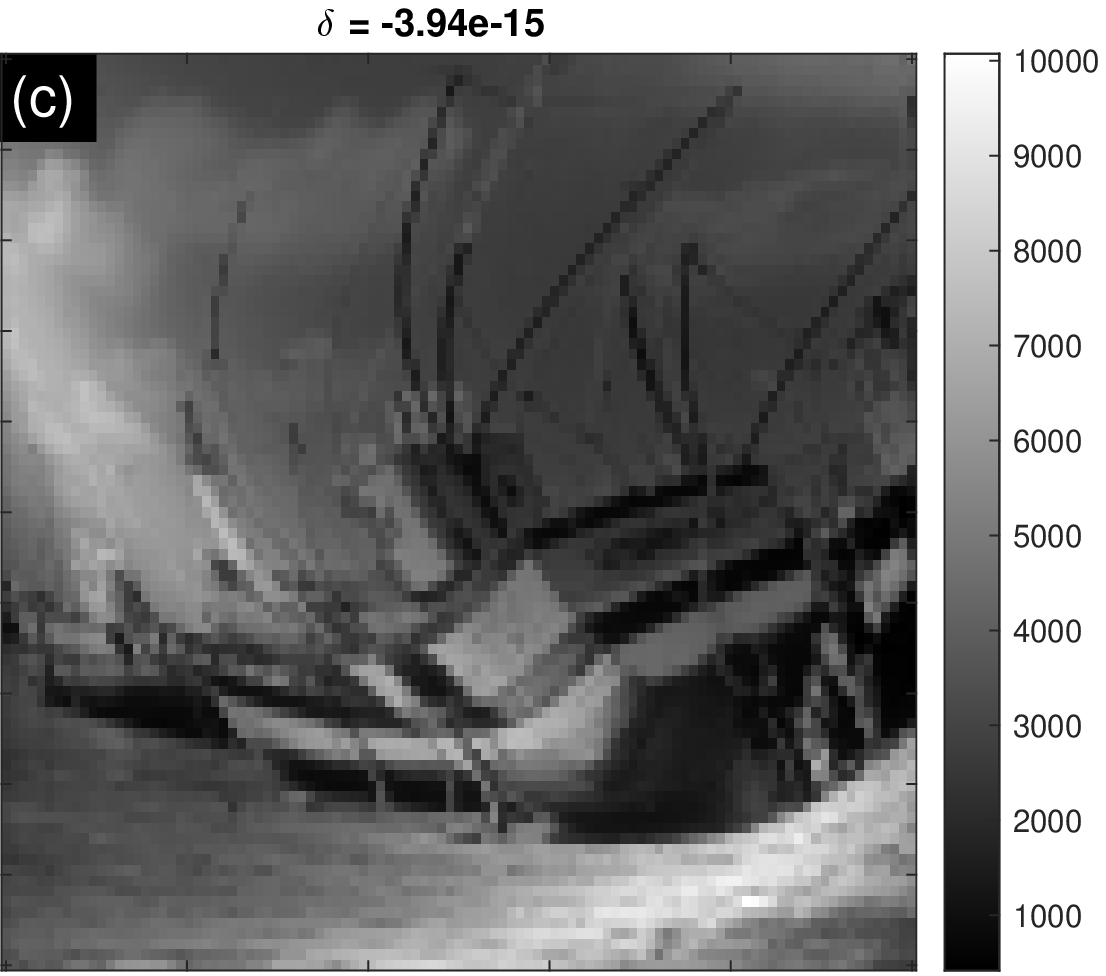}\includegraphics[height=6cm]{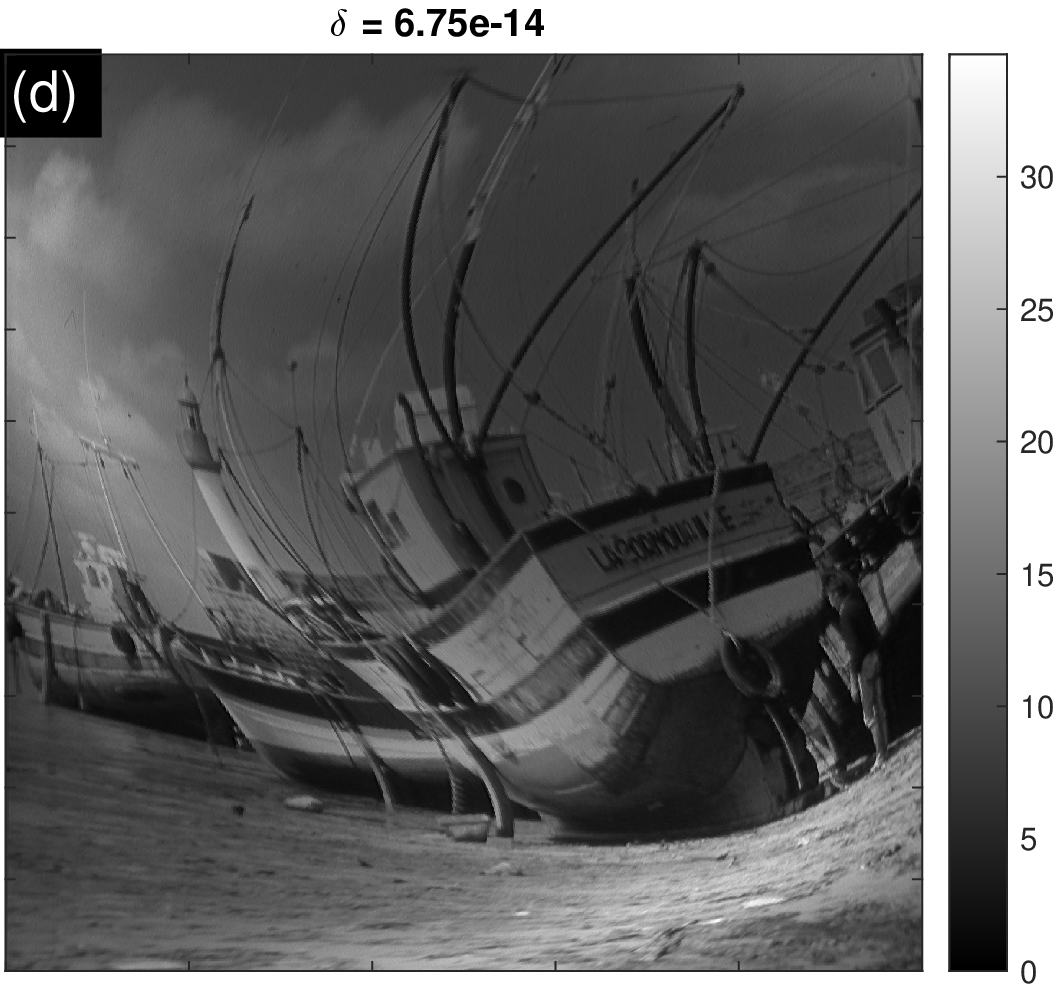}
\par\end{centering}
\caption{Warped boat at different resolutions: a) original image, 512$\times$512;
b) 41$\times$36; c) 105$\times$87, d) 1757$\times$1876. The values
of $\delta$ are reported over each warped image. Color bars at the
side of each panel show how the intensity range of the image is inversely
proportional to its resolution, so that the sum of the pixel values
remains constant.\label{fig:Warped-boat}}
\end{figure}

The result of warping is shown in Figure \ref{fig:Warped-boat}. The
warp has been computed at various completely arbitrary resolutions,
under and oversampling the image, to show the generality of the procedure.
The calculation is performed in double precision floating point. The
resulting $\delta$, reported over each warped image, are barely over
numerical precision.%

\section{Extensions\label{sec:Extensions}}

\begin{figure}
\begin{centering}
\includegraphics[width=0.25\textwidth]{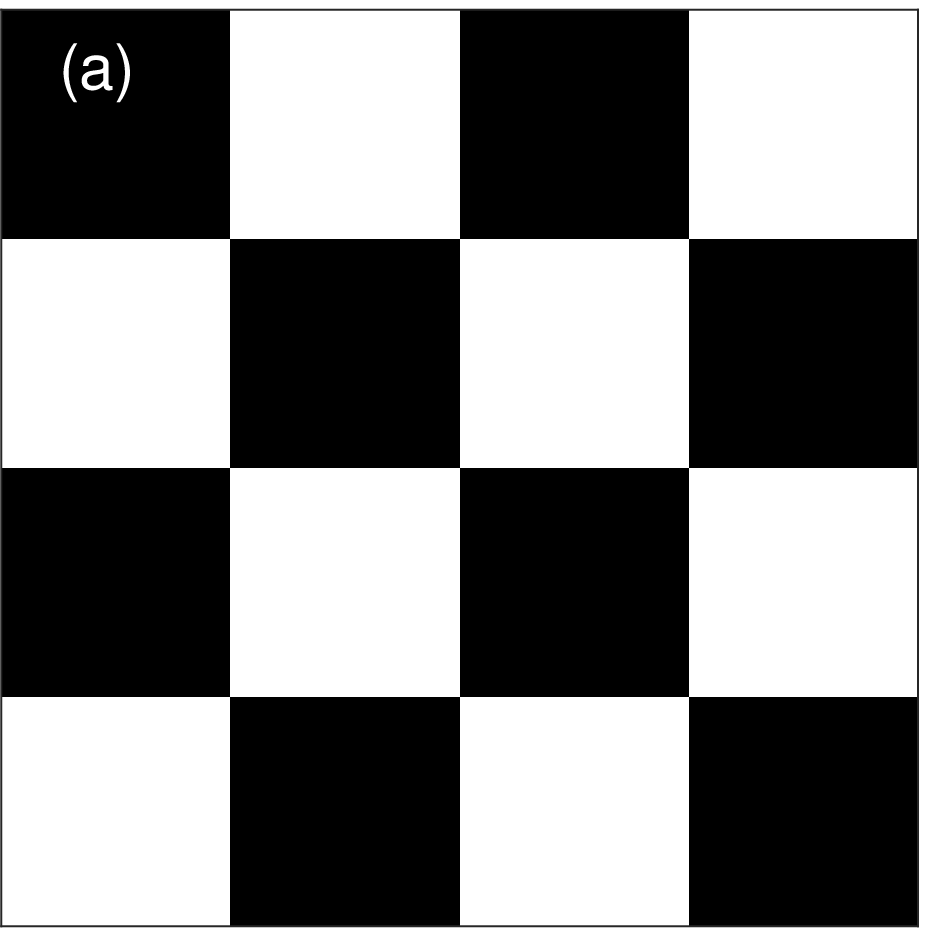}~\includegraphics[width=0.25\textwidth]{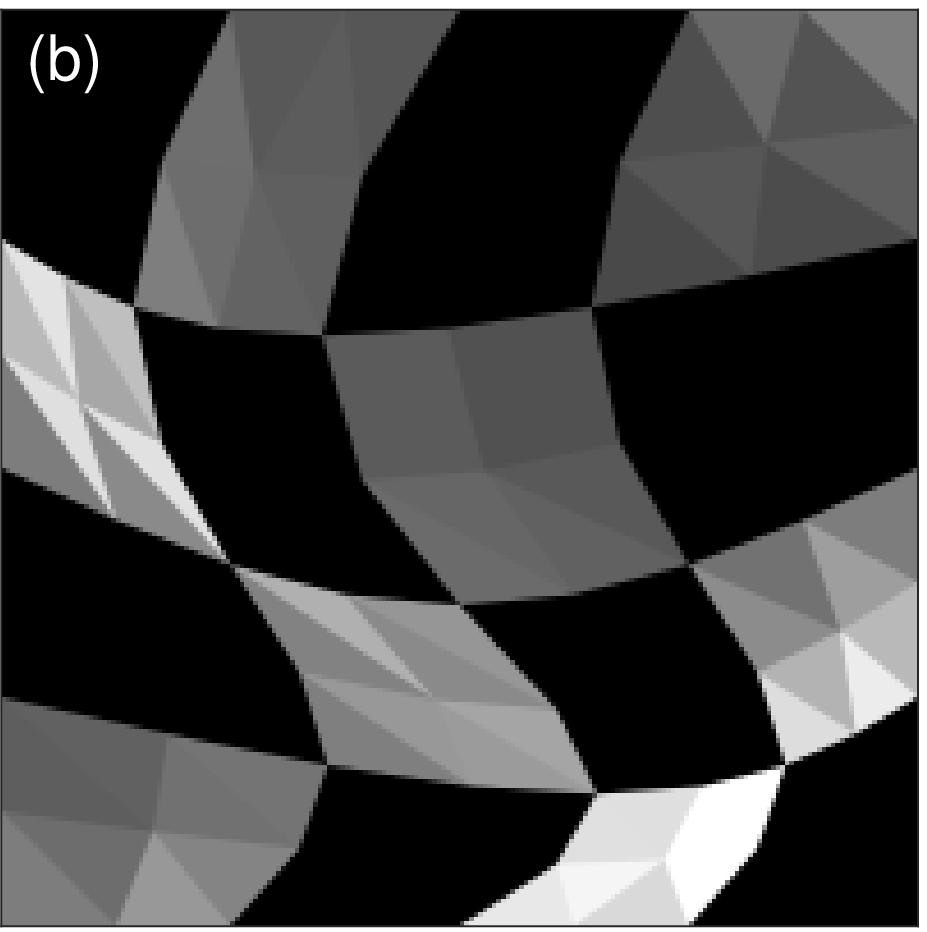}~\includegraphics[width=0.25\textwidth]{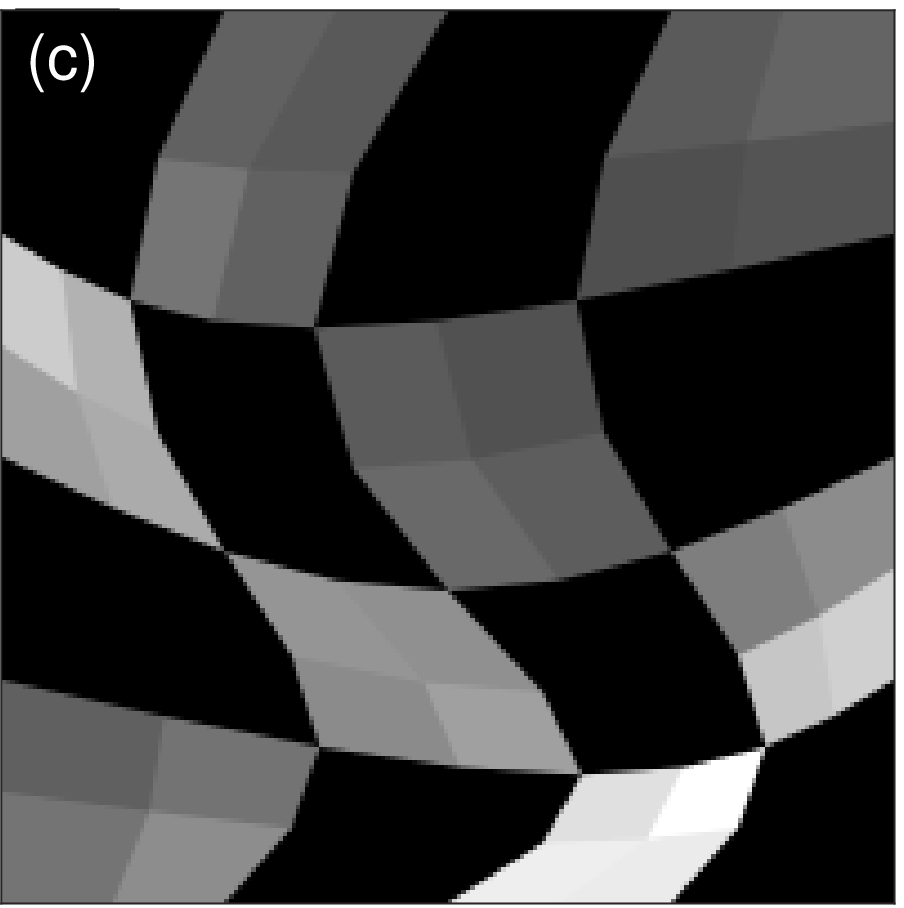}
\par\end{centering}
\caption{Effect of hemipixel vs.~full pixel weighting on a highly oversampled
transformation. a) Original $8\times8$ pixels image, in which each
square of the checker pattern occupies $2\times2$ pixels. b) Warp
to $200\times200$ pixels using Eq.~(\ref{eq:intensity-triangle-contributions}).
The different brightness of the halves of highly stretched pixels
is apparent. c) Warp to $200\times200$ pixels using Eq.~(\ref{eq:quadrangle_contribution}),
which averages the contributions of the two original halves. The grayscale
map of a) is different than that of b) and c) in order to stretch
the visual contrast.\label{fig:Effect-of-hemipixel}}
\end{figure}

Alternative forms of the matrix element $B_{lm,ij}$ can be devised,
giving different weights to the deformed pixel overlaps. The form
of (\ref{eq:intensity-triangle-contributions}) distributes the available
intensity separately on each destination hemipixel. Its effect can
be appreciated in figure (\ref{fig:Effect-of-hemipixel}) for high
downsampling ratio and non-affine pixel deformations, for which $\mathcal{A}\left[f\left(t_{ij}^{L}\right)\right]$
is significantly different from $\mathcal{A}\left[f\left(t_{ij}^{U}\right)\right]$.
Two other choices are presented in the following.

\subsection{Pixel uniformity}

Grouping together the two hemipixels, we can recast the total intensity
of the original rectangular pixel onto the destination pixel, as actually
prescribed by Eq.~(\ref{eq:pixel_overlap_contributions}): 

\begin{equation}
B_{lm,ij}^{q}=\frac{\mathcal{A}\left[T_{lm}^{U}\cap f\left(t_{ij}^{U}\right)\right]+\mathcal{A}\left[T_{lm}^{L}\cap f\left(t_{ij}^{U}\right)\right]+\mathcal{A}\left[T_{lm}^{U}\cap f\left(t_{ij}^{L}\right)\right]+\mathcal{A}\left[T_{lm}^{L}\cap f\left(t_{ij}^{L}\right)\right]}{\mathcal{A}\left[f\left(t_{ij}^{U}\right)\right]+\mathcal{A}\left[f\left(t_{ij}^{L}\right)\right]}\,.\label{eq:quadrangle_contribution}
\end{equation}
In other words, the contributions of the transformed half pixels $f\left(t_{ij}^{U}\right)$
and $f\left(t_{ij}^{L}\right)$ are weighted with a cumulative factor,
which is the average of the two denominators in (\ref{eq:intensity-triangle-contributions}).
The effect of this choice is illustrated in Fig.~\ref{fig:Effect-of-hemipixel}.
It may be argued that this weighting is more natural, as it preserves
the original quadrangular pixel identity, rather than splitting it
arbitrarily along one of its two diagonals.

\subsection{Weighted area interpolation}

In the intensity-preserving resampling illustrated before, we assign
a contribution of pixel $I_{1}(i,j)$ to the destination pixel $P_{lm}$
which is proportional to to the inverse area of $Q_{ij}$, i.e. we
take into account the local stretch or contraction caused by the warp.
If instead we weight the contribution according to the destination
area covered, normalizing over the area of the destination pixel,
as in

\begin{equation}
B_{lm,ij}^{a}=\frac{\mathcal{A}\left[T_{lm}^{U}\cap f\left(t_{ij}^{U}\right)\right]+\mathcal{A}\left[T_{lm}^{L}\cap f\left(t_{ij}^{U}\right)\right]+\mathcal{A}\left[T_{lm}^{U}\cap f\left(t_{ij}^{L}\right)\right]+\mathcal{A}\left[T_{lm}^{L}\cap f\left(t_{ij}^{L}\right)\right]}{\mathcal{A}\left[T_{lm}^{L}\right]+\mathcal{A}\left[T_{lm}^{U}\right]}\,,\label{eq:destination-equalization}
\end{equation}
we achieve a form of area averaging and resampling. For a Cartesian
grid, obviously $\mathcal{A}\left[T_{lm}^{L}\right]+\mathcal{A}\left[T_{lm}^{U}\right]=\Delta X\Delta Y$.
When undersampling, i.e. when several transformed pixels fall into
a single destination pixel, their intensity values are averaged with
a weight proportional to the area which they occupy on the destination;
when oversampling, i.e. as a single deformed pixel covers more than
a destination pixel, that destination pixel is assigned the same intensity
of the source. This form can be thus seen as a value preserving warping,
rather than an intensity preserving warping, and can be compared to
other interpolation techniques in use in image processing.

\subsubsection{Comparison with resampling interpolation\label{subsec:Comparison-with-resampling}}

\begin{figure}
\begin{centering}
\includegraphics[width=0.4\textwidth]{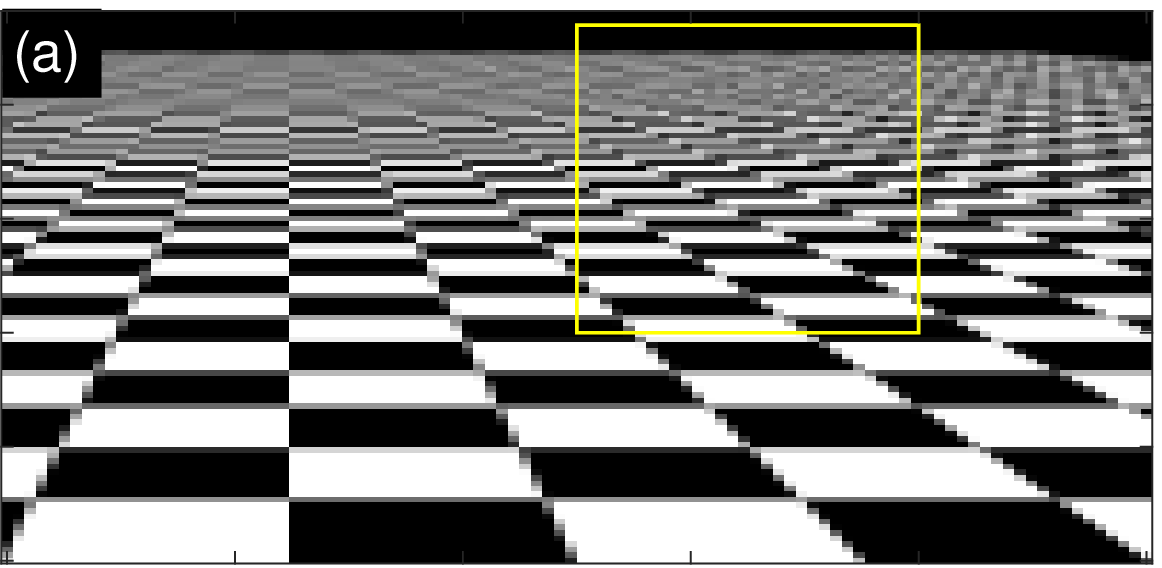}~\includegraphics[width=0.4\textwidth]{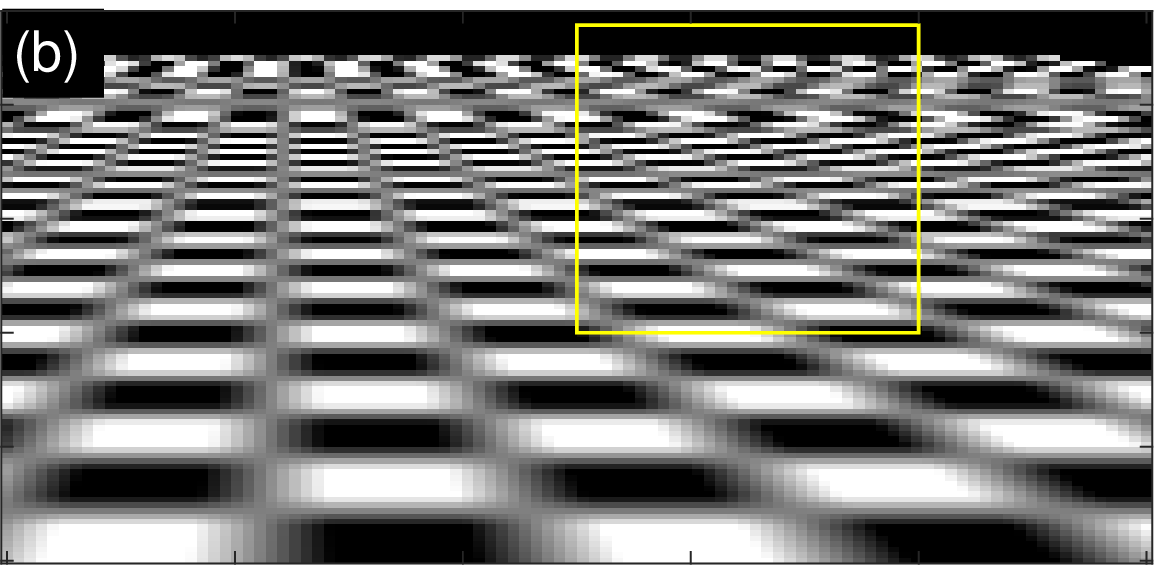}
\par\end{centering}
\begin{centering}
\includegraphics[width=0.3\textwidth]{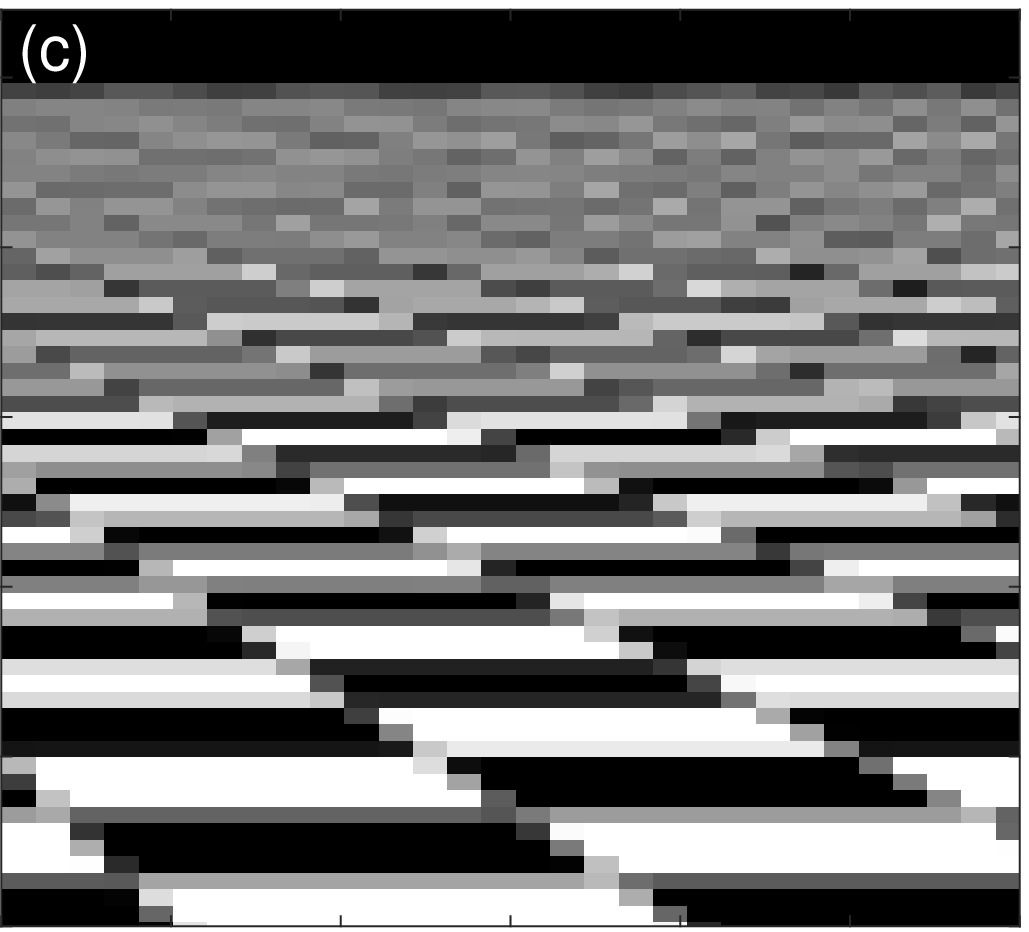}\hspace{0.06\textwidth}\includegraphics[width=0.3\textwidth]{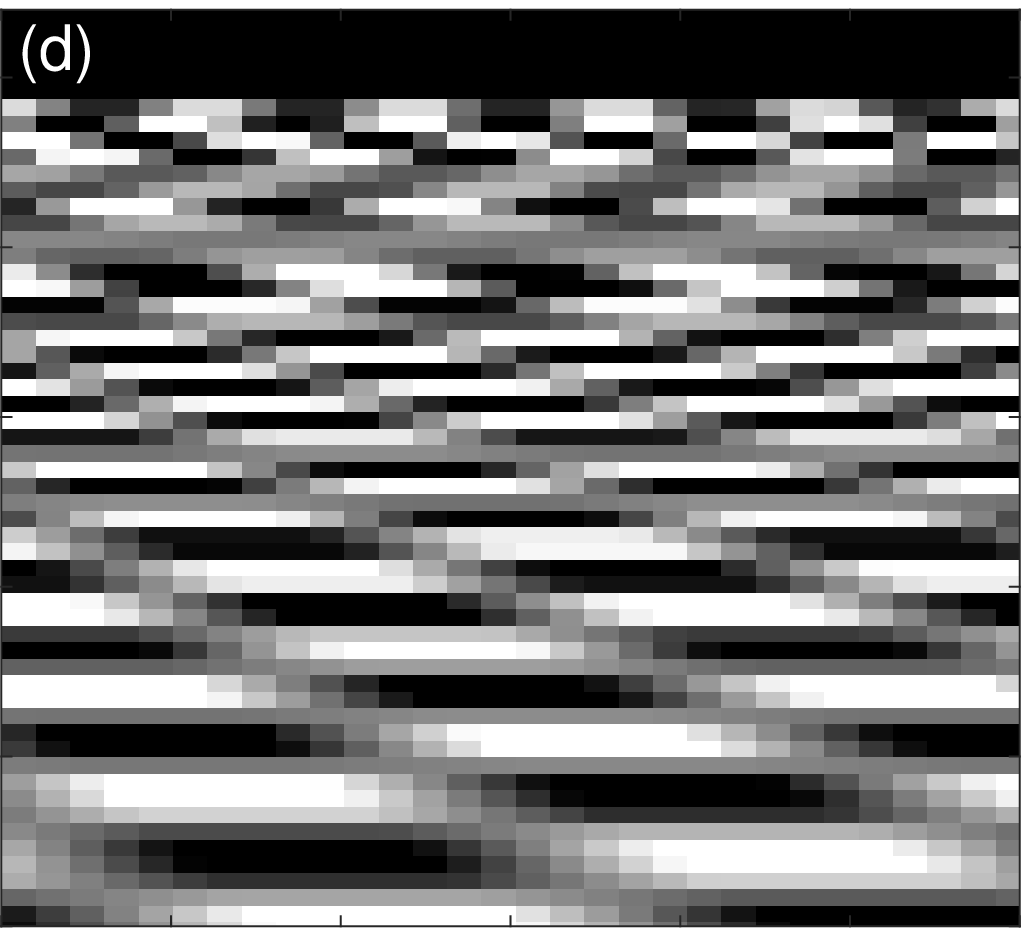}
\par\end{centering}
\caption{Comparison between area resampling (left) and unfiltered bilinear
interpolation resampling (right). The original image is a checker
pattern defined on $128\times64$ pixels, which is projected with
(\ref{eq:perspective}-\ref{eq:inverse-perspective}) onto a raster
of $100\times100$ rectangular pixels. The lower panels (c) and (d)
show a zoom-in of the area enclosed within the yellow boxes. Differences
are appreciated as absence of smoothing at the checker boundaries
on the ``near'' side of the checker, and as reduction of aliasing
artifacts at the ``far'' end.\label{fig:Comparison-resampling}}

\end{figure}

Figure \ref{fig:Comparison-resampling} provides a visual comparison
of the merits of the weighted area interpolation based on Eq.~(\ref{eq:destination-equalization})
versus the commonly used bilinear interpolation, as a reference. Other
more sophisticated interpolators, like higher order polynomial (e.g.
bicubic, spline), Lanczos, or edge preserving (Akima), etc.~could
be compared as well, without affecting the main result. While many
more interpolation methods are known in literature, the comparison
with the simplest baseline algorithm is justified by the fact that
the present area resampling recipe is only dependent on the geometry
of the coordinate transform, not on the image data itself, nor on
any assumed or a priori knowledge about the structure of the image.
Other data-dependent interpolators (e.g. Takeda's kernel regression
\cite{Takeda2007}, not to mention even more elaborate techniques
based on deep learning) may produce more ``realistic'' results on
the perceptual point of view, or even behave well as image denoisers
(which implies a discrimination between an underlying image model
and the superimposed corrupting noise). Here we merely report about
the own merits of the weighted area resampler, without claiming that
it is outperforming other image reconstruction techniques.

As an example of warping, we use the perspective transformation
\begin{equation}
\left(\begin{array}{c}
X\\
Y
\end{array}\right)=g(x,y)=\left(\begin{array}{c}
a+\left(x-a\right)\frac{d-b}{y-b}\\
c\left(1+\frac{d-b}{y-b}\right)
\end{array}\right)\,,\label{eq:perspective}
\end{equation}
which describes the projection of an image on the $xy$ plane on the
vertical plane $x=X$, $y=d$, $Y=z$, from the viewpoint $x=a$,
$y=b$, $z=c$.

To compare the two, we apply (\ref{eq:matrix-intensity-transformation})
and (\ref{eq:destination-equalization}) using areas of triangles
transformed from the $(x,y)$ to the $(X,Y)$ space, i.e. using a
\emph{direct} coordinate transform. For the interpolation instead,
we exploit a more efficient, customary implementation which evaluates
the image values on the regular $(X,Y)$ destination grid by looking
up and interpolating values on $g^{-1}(X,Y)$. In other words, we
perform an \emph{inverse} pixel lookup. The transformation (\ref{eq:perspective})
has analytical inverse
\begin{equation}
\left(\begin{array}{c}
x\\
y
\end{array}\right)=g^{-1}(X,Y)=\left(\begin{array}{c}
\frac{2ac-cX-aY}{c-Y}\\
\frac{bY+ce-2bc}{c-Y}
\end{array}\right)\,,\label{eq:inverse-perspective}
\end{equation}
with $a=\frac{1}{4}$, $b=-\frac{1}{10}$, $c=\frac{1}{2}$, $d=0$,
and $-5<J_{g}<-\frac{5}{1331}$ for $0<y<1$. To stress the essential
differences between the two methods, no dealiasing filter prior to
interpolation is applied.

\begin{figure}
\begin{centering}
\includegraphics[height=0.35\columnwidth]{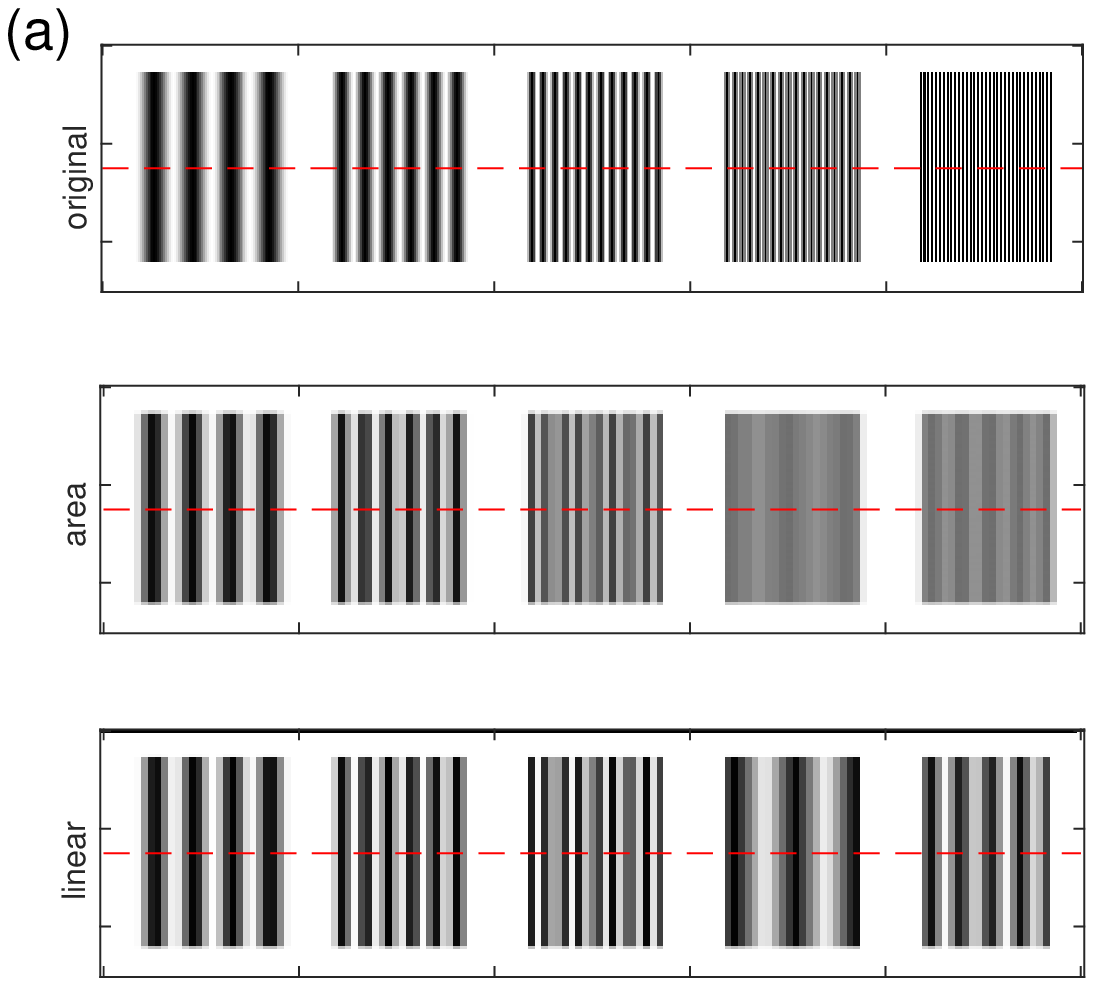}\qquad{}\includegraphics[height=0.35\columnwidth]{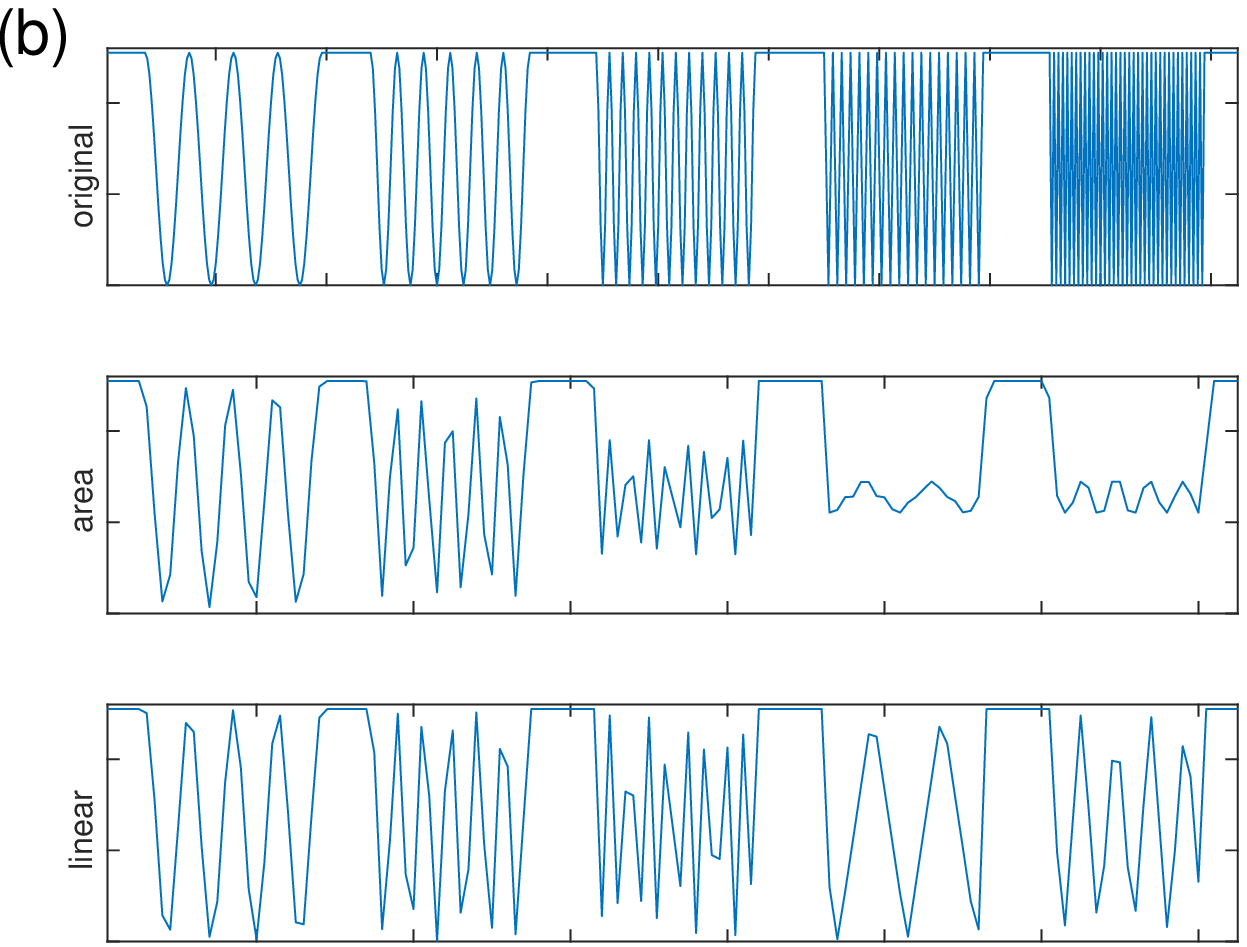}
\par\end{centering}
\caption{(a) A $512\times100$, 8 bit image containing sinusoidal test bars,
with periods of 20, 12, 6, 4 and 2 pixels, (top) downsampled to $145\times80$
using area resampling (center) and linear interpolation (the pixel
aspect ratio is varied so that all images have the same coordinate
span). (b) intensity profiles of the three images along the center
line, red dashed lines in (a). For these bar spatial frequencies and
resampling ratio, the amplitude of the aliased spatial frequency of
the rightmost patterns is reduced using area downsampling. \label{fig:bars-aliasing}}
\end{figure}

Area resampling somehow averages many original image values falling
on the destination pixels, whereas the traditional resampling only
picks up one value or averages few neighbors, sampling those which
fall close to an interpolation point. Area resampling provides thus
a smoother result than interpolation when undersampling images, and
can be less prone to aliasing, since in that case it automatically
behaves as an adaptive box filter. Interpolation is affected by aliasing,
which is usually cured by low pass prefiltering; however, for a general
warp transformation this filtering has to be local \cite{Heckbert1989},
complicating matters. An example of aliasing reduction is shown in
Figure \ref{fig:bars-aliasing}, where a global rescaling ratio and
a simple periodic pattern evidence the different amplitude of the
aliased spatial component. The amount of alias suppression, though,
is entirely dependent on the particular local downsampling ratio,
and on the original image content.

On the other hand, when oversampling, the weighted area transform
produces images which are sharper and more faithful to pixel edges,
since the destination value picked in that case represents well the
original pixel value, rather than being an interpolation between nearest
neighbors.

\section{Example: source photometry\label{sec:Example:-source-photometry}}

\begin{figure}
\begin{centering}
\includegraphics[height=0.15\paperheight]{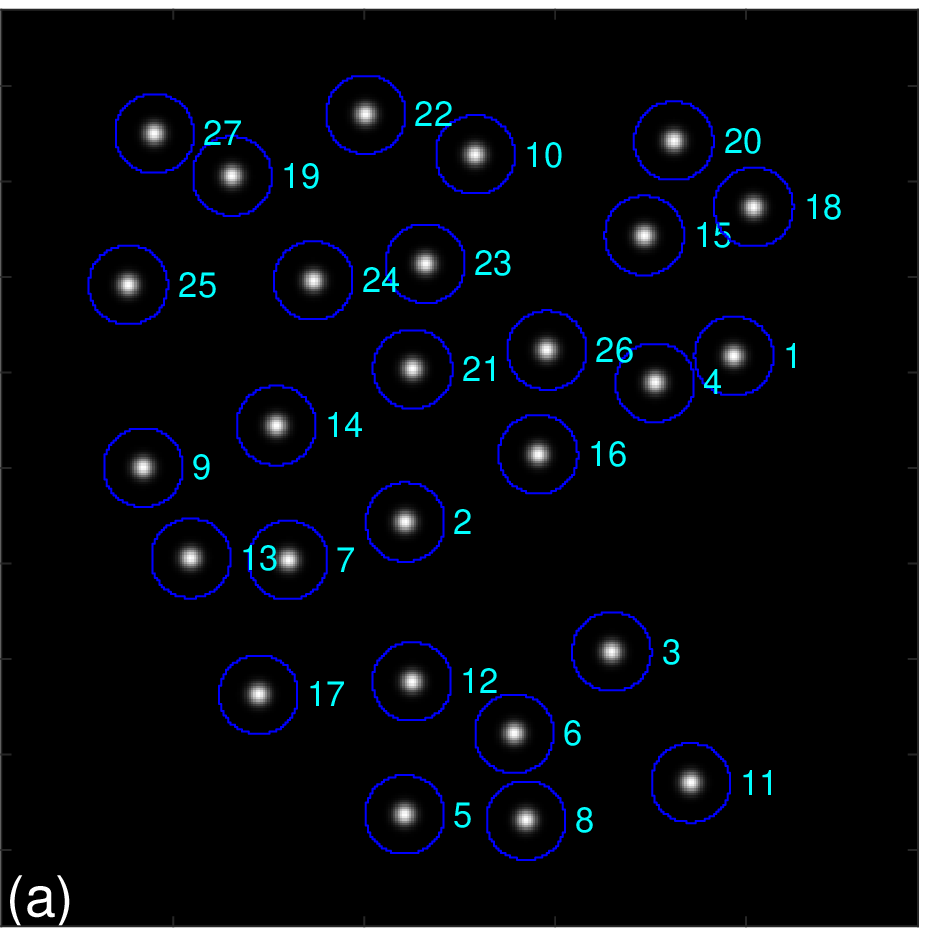}~~\includegraphics[height=0.15\paperheight]{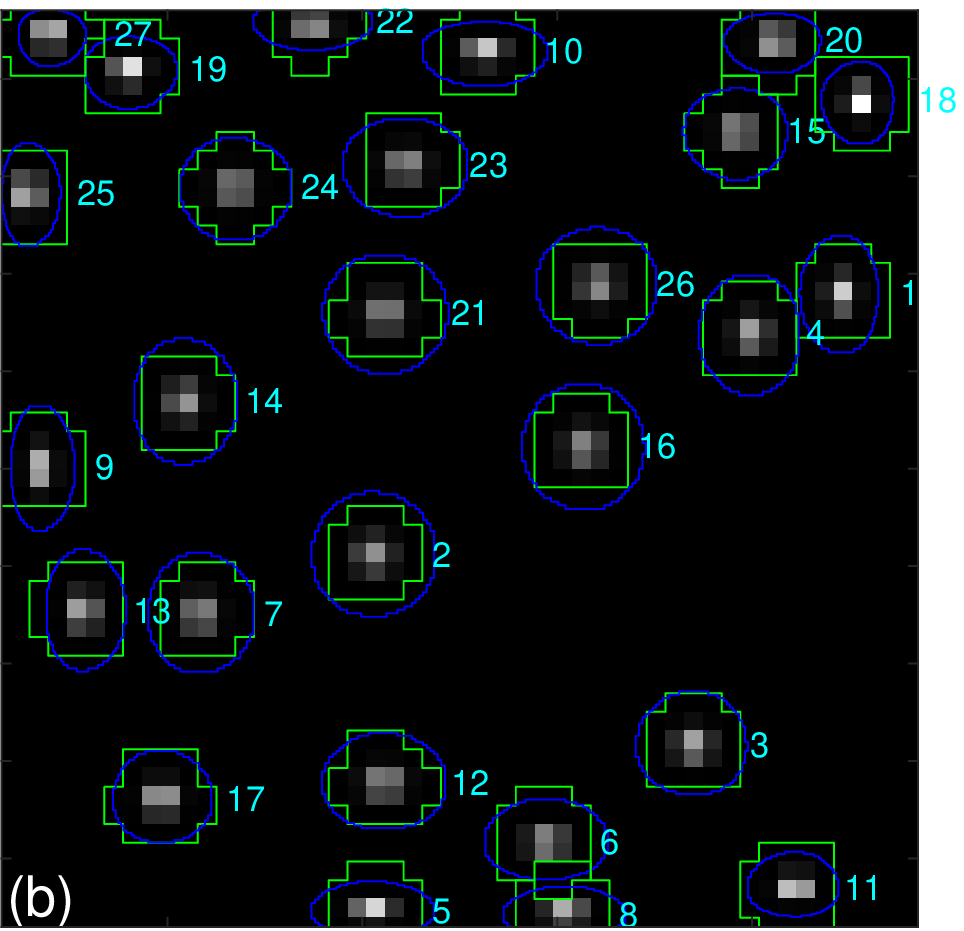}
\par\end{centering}
\begin{centering}
\includegraphics[height=0.15\paperheight]{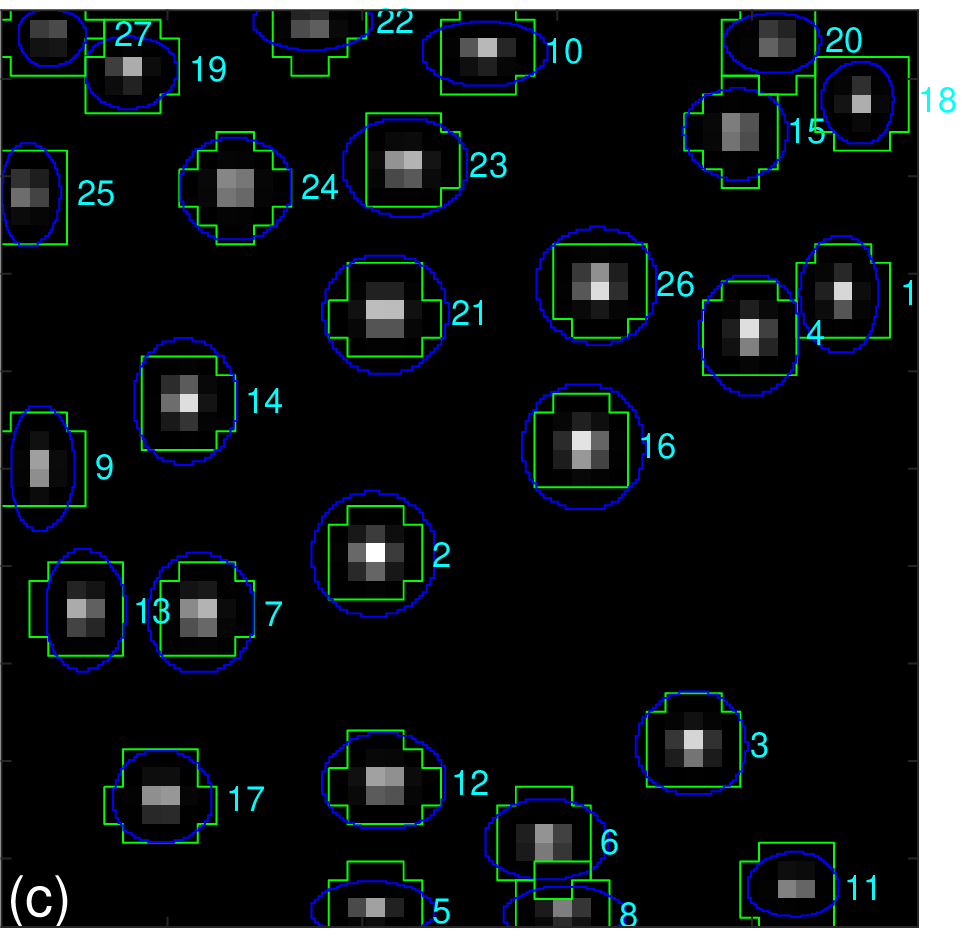}~\includegraphics[height=0.15\paperheight]{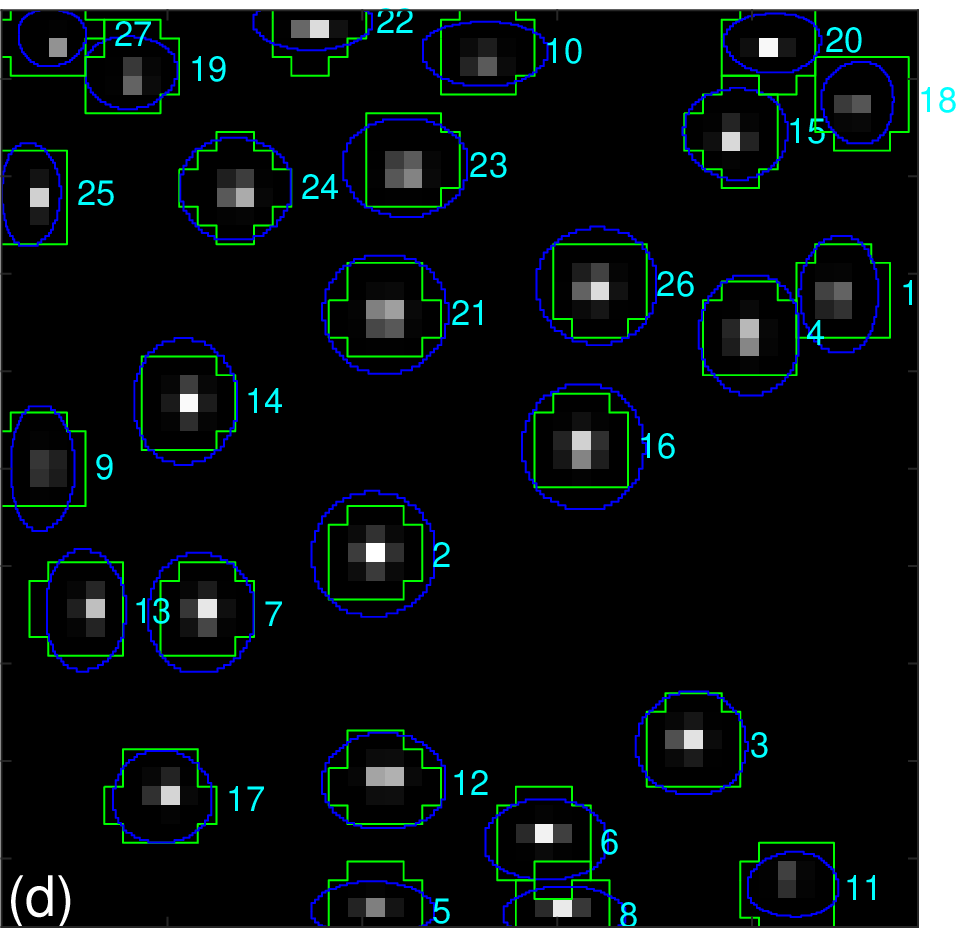}
\par\end{centering}
\caption{Photometry of synthetic sources in a warped image: a) original image
$I_{1}$, 400$\times$400. The intensity $\tilde{s}_{k}^{\text{orig}}$
of each dot is evaluated summing up all pixel values in neighborhoods
of radius $4\sigma$, whose contour is plotted in blue. b) Image $I_{2}$,
warped and downsampled to 50$\times$50 using Eq.~(\ref{eq:quadrangle_contribution}).
New pixel neighborhoods of radius $4\sigma$ around the displaced
source centers are plotted in green, along with the deformed contours
of the original neighborhoods, in blue. c) Image $I_{2}^{w}$, warped
to 50$\times$50 using the weight of Eq.~(\ref{eq:destination-equalization}).
d) $I_{2}^{\text{interp}}$, warped resampling to 50$\times$50 of
$I_{1}$, using bilinear interpolation.\label{fig:Warped-stars}}
\end{figure}

Preservation of the intensity is an essential property when photometry
is performed on deformable images. We show in figure \ref{fig:Warped-stars}
an illustrative example with synthetic data. To make our point, we
consider the warp of a high resolution image to a lower resolution,
and different ways to estimate, in the transformed image, the original
intensity of each source. An original $400\times400$ pixels image
is created, simulating well separated, randomly placed sources with
a gaussian peak profile with $\sigma=0.01L$, where $L=1$ is the
size of the square image. Each source $k$, centered at $\left(x_{k}^{s},y_{k}^{s}\right)\in]0,1[\otimes]0,1[$,
contributes to the pixel $I_{1}(i,j)$ with intensity
\begin{equation}
I_{1}^{k}(i,j)=\frac{1}{4}\left[\text{erf}\left(\frac{x_{i}-x_{k}^{s}+\Delta x}{\sigma}\right)-\text{erf}\left(\frac{x_{i}-x_{k}^{s}}{\sigma}\right)\right]\,\left[\text{erf}\left(\frac{y_{j}-y_{k}^{s}+\Delta y}{\sigma}\right)-\text{erf}\left(\frac{y_{j}-y_{k}^{s}}{\sigma}\right)\right]\,.\label{eq:pixel-gaussian}
\end{equation}
With this integral definition, the total contribution of each source
is normalized to the value $s_{k}=1$. The image is then warped and
downsampled to $50\times50$ using the transformation
\begin{equation}
\left(\begin{array}{c}
X\\
Y
\end{array}\right)=f(x,y)=\left(\begin{array}{c}
\frac{1-\cos\left(\pi x\right)}{2}\\
\frac{1-\cos\left(\pi y\right)}{2}
\end{array}\right)\label{eq:sin-mapping}
\end{equation}

which expands the original image away from its center, compressing
it at the edges. Differently than (\ref{eq:wavy}), this mapping has
a closed inverse form
\begin{equation}
\left(\begin{array}{c}
x\\
y
\end{array}\right)=f^{-1}(X,Y)=\left(\begin{array}{c}
\frac{1}{2}-\frac{\sin^{-1}\left(1-2X\right)}{\pi}\\
\frac{1}{2}-\frac{\sin^{-1}\left(1-2Y\right)}{\pi}
\end{array}\right)\,.\label{eq:arcsin-mapping}
\end{equation}

The Jacobian of the direct transformation is $J_{f}=\frac{1}{4}\pi^{2}\sin(\pi x)\sin(\pi y)$,
whereas that of its inverse is 
\[
J_{f}^{-1}=\frac{1}{\pi^{2}\sqrt{X\left(X-1\right)Y\left(Y-1\right)}}\,.
\]
While this mapping is not exceedingly representative of the transformations
used in practice to correct imaging defects (which are often modeled
by polynomial functions), the existence of an analytical inversion
formula instead of an approximation to it, allows a fair comparison
with traditional resampling image interpolations, which are easily
performed using the inverse map.

\begin{figure}
\noindent \begin{centering}
\includegraphics[width=0.47\textwidth]{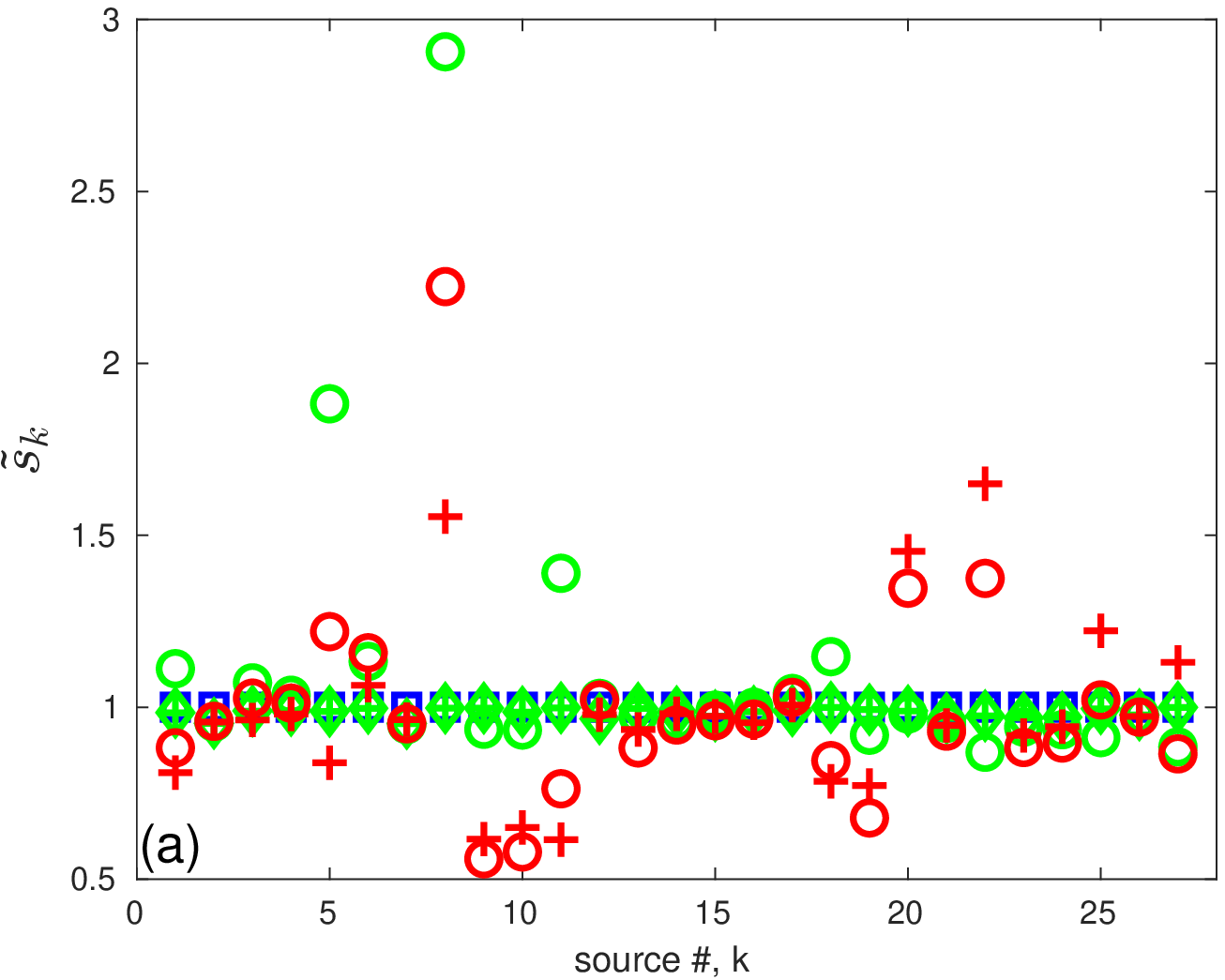}\includegraphics[width=0.48\textwidth]{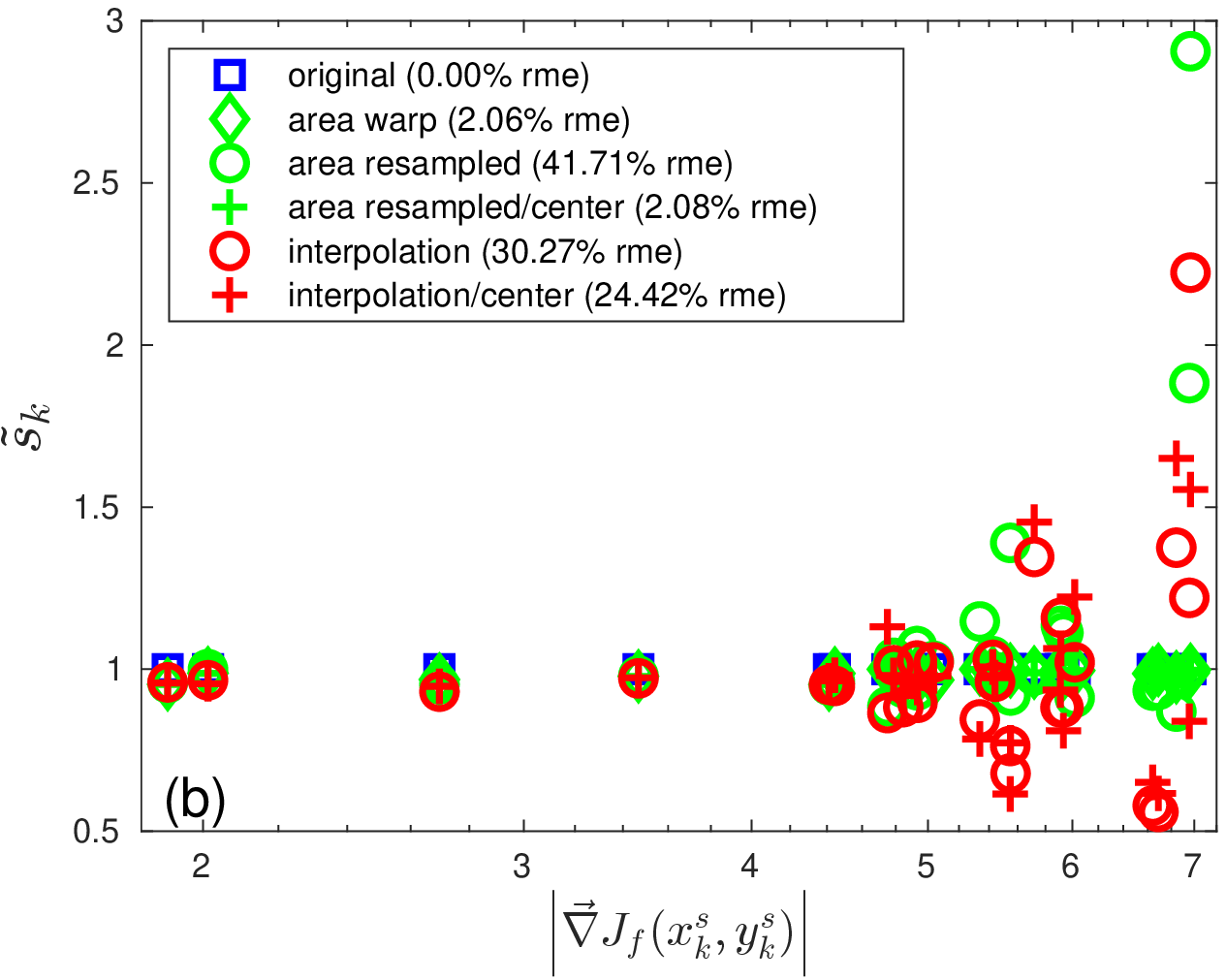}
\par\end{centering}
\caption{Source intensities $\tilde{s}_{k}$ estimated according to different
methods. The root mean square error of each method is reported in
the caption. In (a), the datapoints are sorted according to the source
number $k$, for identification on Figure \ref{fig:Warped-stars};
in (b) the inferred source strengths are plotted versus the value
of the gradient of the Jacobian at the center of the source. The plot
shows how, while area warp provides a reasonable photometric estimation
everywhere in the warped domain, methods 5 and 6 succeed only where
$J_{f}$ varies slowly. The relative better performance of method
6 over 5 is probably due to the symmetry of the original source, and
the better representativeness of the central value of $J_{f}$.\label{fig:reconstructed-peak-intensities}}

\end{figure}

To estimate a posteriori the contribution of the sources, all intensity
values in a pixel neighborhood of radius $4\sigma$ are summed, and
the result is compared to the nominal unit intensity of the peak.
To simulate possible pitfalls of the process, when applied to real
images, we add some real world methodological errors. In figure \ref{fig:Warped-stars}a,
we compute the pixel sums even when some of the neighborhoods receive
overlapped contributions from more than a source, or  sources fall
near the margins of the image and contribute incompletely to the total.
In figure \ref{fig:Warped-stars}b, we estimate the intensity of each
source by summing pixel values within a circular pixel neighborhood
centered on the transformed source position, rather than transforming
the shape of the initial neighborhood. In figure \ref{fig:Warped-stars}c
and d, we interpolate the image, compute the neighborhood sums, compensating
for the area changes by either multiplying them by $J_{f}^{-1}\left(x_{k}^{s},y_{k}^{s}\right)$
computed merely at the source center, or by multiplying the local
intensity by the local value of the Jacobian. In total we compare
six different estimators of the intensity of the source, assuming
$\left(x_{k}^{s},y_{k}^{s}\right)$ known a priori:
\begin{enumerate}
\item On the original image, we compute $\tilde{s}_{k}^{\text{orig}}=\sum_{4\sigma}I_{1}(i,j)$
over the pixels within a distance $4\sigma$ from the center $\left(x_{k}^{s},y_{k}^{s}\right)$
(within the blue contours in figure \ref{fig:Warped-stars}a)
\item On the area warped image, we compute $\tilde{s}_{k}^{\text{area\,warp}}=\sum_{4\sigma}I_{2}(l,m),$
summing this time the intensities of the destination pixels which
fall within $4\sigma$ from the \emph{transformed} center $\left(X_{k}^{s},Y_{k}^{s}\right)$
(within the green contours in figure \ref{fig:Warped-stars}b). These
summation neighborhoods may differ form the transformed original ones
(compare green and blue lines); by using this evaluation, we want
to assess the error involved, which is presumed small given the rapid
decay and good separation of the peaks.
\item Using instead the area resampled image $I_{2}^{w}$ of figure \ref{fig:Warped-stars}c,
obtained using the weighted equalization of (\ref{eq:destination-equalization}),
we compute 
\[
\tilde{s}_{k}^{\text{area\,resampled}}=\frac{\Delta X\Delta Y}{\Delta y\Delta y}\,\sum_{4\sigma}J_{f}^{-1}(l,m)\cdot I_{2}^{w}(l,m)
\]
Here $J_{f}(l,m)$ is the value of the Jacobian evaluated at the center
of each pixel of the destination image.
\item Using the area resampled image of figure \ref{fig:Warped-stars}c,
we compute 
\[
\tilde{s}_{k}^{\text{area\,resampled/center}}=J_{f}^{-1}\left(X_{k}^{s},Y_{k}^{s}\right)\,\frac{\Delta X\Delta Y}{\Delta y\Delta y}\,\sum_{4\sigma}I_{2}^{w}(l,m)
\]
 using the value of the Jacobian evaluated at the transformed position
of the center of the source alone.
\item Using instead the warped and interpolated image $I_{2}^{\text{interp}}(l,m)$
of figure \ref{fig:Warped-stars}d, we compute 
\[
\tilde{s}_{k}^{\text{interpolation}}=\frac{\Delta X\Delta Y}{\Delta y\Delta y}\,\sum_{4\sigma}J_{f}^{-1}(l,m)\cdot I_{2}^{\text{interp}}(l,m)
\]
\item Using the interpolated image $I_{2}^{\text{interp}}(l,m)$, we compute
\[
\tilde{s}_{k}^{\text{interpolation/center}}=J_{f}^{-1}\left(X_{k}^{s},Y_{k}^{s}\right)\,\frac{\Delta X\Delta Y}{\Delta y\Delta y}\,\sum_{4\sigma}I_{2}^{\text{interp}}(l,m)
\]
\end{enumerate}
Figure \ref{fig:reconstructed-peak-intensities} summarizes the results
of the various estimators. Discrepancies of the different $\tilde{s}_{k}$
with respect to the nominal value can be ascribed to sums over circular
neighborhoods which deviate from the actual warped profile of the
peak, as well as to overlapping peak tails (which are minimal in our
example), but are notably due to the use of Jacobian factors evaluated
pointwise. Figure \ref{fig:reconstructed-peak-intensities}b shows
in particular how errors significantly increase for sources at the
periphery of the domain, where $J$ varies more rapidly, as quantified
by $\left|\vec{\nabla}J_{f}\right|$, which is easily computed analitically.
We have chosen here deliberately an extreme case, in which $J$ is
not constant across the domain, and a downsampling of a factor 8,
in order to exhacerbate the loss of information due to point-based
resampling. It is clear that naive estimators based on the interpolated
image can lead to misestimation, with root mean square errors $\varepsilon=\sqrt{\left\langle \left(\tilde{s}_{k}-s_{k}\right)^{2}\right\rangle }$
of the order of a quarter of the nominal peak intensity itself. In
contrast, even a naive estimation, using mere undeformed circular
neighborhoods, on the area warped image ($\tilde{s}_{k}^{\text{area\,warp}}$
of point 2) produces much more accurate results.

\section{Conclusions and future outlook\label{sec:Conclusions-and-future}}

We have described a rigorous method for warping images, which by construction
preserves the cumulative brightness of their features, and thus is
suitable for photometric measurements on the transformed image. In
doing so we dwelt on the computational geometry problem of finding
the area of the intersection of two triangles. Despite its geometrical
simplicity, we were not aware of a viable and robust algorithm for
it available openly, and we provide one. We showed that a slight variation
of the procedure can instead preserve the local values of intensity,
and thus be directly compared with traditional implementations of
warping, based on resampling the deformed image at gridpoints. Our
method remains an area resampling method also in this application,
and thus has implicit different filtering properties, not requiring
for instance a preliminary antialiasing filter, and preserving sharper
edges in case of severe oversampling.

The method has been showcased on monochrome images, but its extension
to multichannel (e.g.~color) images would be trivial, and in its
simplest conception would amount to the computation of equation (\ref{eq:matrix-intensity-transformation})
independently for each channel.

The algorithms proposed are computationally more demanding that plain
resampling ones, and in this work we have not pursued their highest
possible efficiency. Future work could concentrate on developing faster
implementations of them. Being prone to parallelization (see Appendix
\ref{app:Computational-performance}), a GPU implementation of the
algorithm can be envisioned. Once proved viable, the implementation
of the present method in different programming languages its and inclusion
into popular software packages like those mentioned in the introduction
can be advocated for.

As a further development, the procedure could be adapted to non-rectangular
source pixels, which can be in any case be decomposed into constituent
triangles. The core of the method would remain the same, the only
differences would be in devising an indexing for the triangulation
of the shaped pixels. Two use cases come in mind: for one, real physical
imagers, notably CMOS sensor chips, have by architectural necessity
photosensitive areas which cover partially the rectangular pixel cell
\cite{Yadid-Pecht2004}. Our procedure would provide an area-consistent
way of resampling their measurements on differently gridded or deformed
coordinates. As for another application, our method could be used
where by design the image pixels are not arranged over a Cartesian
grid at all, like for instance in hexagonal image processing \cite{Middleton2005}.

\appendix

\section{Barycentric coordinates\label{app:Barycentric-coordinates}}

Given the triangle $T=\left\{ \vec{u},\vec{v},\vec{w}\right\} $,
defined by the plane coordinates of its three vertices, and a point
$\vec{x}$, we define:
\begin{align}
2A & =2\left[\left(v_{y}-w_{y}\right)\left(u_{x}-w_{x}\right)-\left(v_{x}-w_{x}\right)\left(u_{y}-w_{y}\right)\right]\nonumber \\
s & =\left(\vec{x}-\vec{w}\right)\cdot\left(v_{y}-w_{y},v_{x}-w_{x}\right)\\
t & =\left(\vec{x}-\vec{w}\right)\cdot\left(u_{y}-w_{y},u_{x}-w_{x}\right)\,,\nonumber 
\end{align}
where $A$ is the signed area of the triangle $T$, positive or negative
depending on the clockwise order of the vertices. Our convention is
to define the (unscaled) barycentric coordinates of point $\vec{x}$
as 
\begin{equation}
\vec{b}=\left(\begin{array}{c}
b_{1}\\
b_{2}\\
b_{3}
\end{array}\right)=\text{sign}(A)\left(\begin{array}{c}
s\\
t\\
2A-s-t
\end{array}\right)\,.
\end{equation}

For any point $\vec{x}$ internal to $T$, $0\le b_{k}\le2A$, for
any $k=\left\{ 1,2,3\right\} $.

If one component $b_{k}=0$, the point $\vec{x}$ lies on a side of
$T$. If two components $b_{k}$ are simultaneously zero, the point
$\vec{x}$ is simultaneously on two sides of $T$, i.e.~coincides
with a vertex of the reference triangle. All three components of $\vec{b}$
can be null only for the degenerate case of a triangle with three
coinciding vertices.

The inversion relation giving $\vec{x}$ from its barycentric coordinates
is

\begin{equation}
\vec{x}=\frac{b_{1}\vec{u}+b_{2}\vec{v}+b_{3}\vec{w}}{2\left|A\right|}\,.
\end{equation}

A segment $\left\{ \vec{x_{1}},\vec{x_{2}}\right\} $, whose extremes
have barycentric coordinates $\vec{b}^{1}$ and $\vec{b}^{2}$ with
respect to $T$, intersects the $k$-th side of the triangle $T$
if $b_{k}^{1}$ and $b_{k}^{2}$ have opposite signs. The barycentric
coordinates $\vec{b}^{c}$ of the intersection point are then
\begin{equation}
\vec{b}^{c}=\frac{b_{k}^{2}\vec{b}^{1}+b_{k}^{1}\vec{b}^{2}}{b_{k}^{2}-b_{k}^{1}}\,.\label{eq:barycrossing}
\end{equation}

\section{Area of the intersection of two triangles\label{app:intersection-area}}

The problem of intersecting triangles in two and three dimensions
has received due attention in computer graphics, being fundamental
in a number of applications which involve triangulation of domains,
like collision detection or intersection of triangulated surfaces
\cite{Elsheikh2014}, and indeed literature on it is available (e.g.~\cite{Yamaguchi1985,Guigue2003,Moeller2004,Tropp2006,Sabharwal2015}).
However, in these works at most the conditions for the detection of
planarity and intersection of two triangles are given, but not an
explicit algorithm computing the overlap area of planar triangles,
which we need here. Its derivation is discussed in this appendix.

\subsection{Enumeration of possible cases\label{subsec:Enumeration-of-possible}}

\begin{figure}
\centering{}\includegraphics[width=0.8\columnwidth]{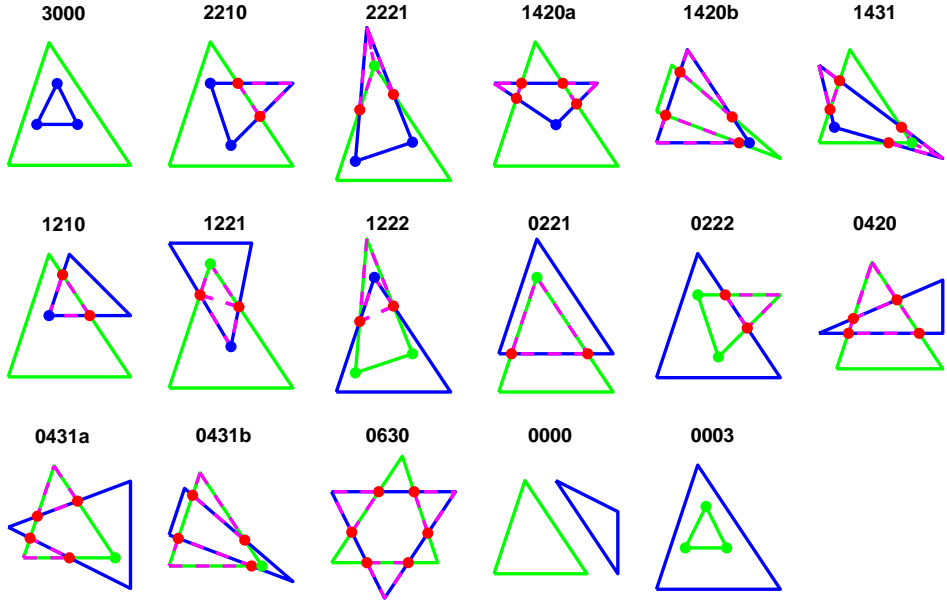}\caption{The seventeen topologically different ways for two triangles $T_{1}$
(blue) and $T_{2}$(green) to intersect. Blue and green dots indicate
respectively vertices of $T_{1}$ internal to $T_{2}$ and viceversa;
red dots intersections between sides. The numeric label above each
couple reflects the classification explained in the text. The area
of the intersection polygon can always be computed as a sum or a difference
of constituent trianglets (up to four), which are highlighted by pink
dashed lines.\label{fig:The-seventeen-topologically}}
\end{figure}

The possible ways in which two triangles $T_{1}$ and $T_{2}$ can
overlap and intersect can be classified according to topological properties.
Being triangles always convex, their intersection is always a convex
polygon. A side of $T_{2}$ can intersect zero, one or two sides of
$T_{1}$. A side of of $T_{2}$ with one vertex internal and one external
to $T_{1}$ implies a single intersection with one side of $T_{1}$,
while both vertices internal or external could both grant either zero
or two intersections. The different overlap cases can be labeled according
to: the number $v_{1}$ of vertices of $T_{1}$ falling inside $T_{2}$
(which can be 0, 1, 2, or 3); the number $i_{s}$ of intersections
between sides of the two triangles (0, 2, 4 or 6); the number $s_{2}$
of sides of $T_{2}$ intersected, and, to remove ambiguities, the
number $v_{2}$ of vertices of $T_{2}$ falling into $T_{1}$. All
the possible cases are depicted in Fig.~\ref{fig:The-seventeen-topologically}.
The label above each subfigure is derived from these numbers, $v_{1}i_{s}s_{2}v_{2}$.
A further distinction is necessary for some of the cases with $i_{s}=4$:
two topologically different arrangements are possible with the same
classification numbers, hence \textbf{1420a} and \textbf{1420b}, \textbf{0431a}
and \textbf{0431b}.

As for areas, clearly $\mathcal{A}\left[T_{1}\cap T_{2}\right]=\mathcal{A}\left[T_{2}\cap T_{1}\right]$,
and commutativity would reduce the number of topologically different
cases to 11 (five of the seventeen cases are topologically invariant
for the exchange of the two members, like e.g.~\textbf{1431},\textbf{
1221}; the other twelve have each one their dual, like \textbf{2210}
and \textbf{0222}, counted only once). In our computation, though,
the two member triangles have different roles, and we must in principle
treat the all cases as distinct. In any case, the area of the intersection
polygon can always be computed as a sum or difference of at most four
smaller trianglets, formed either by the vertices or by the intersection
points of the member sides.

We note that an early analysis of the problem was given in the report
\cite{Sampoli2004}, though without an explicit computation algorithm
and without considering degenerate cases.

\subsection{Degeneracy and numerical precision\label{subsec:Degeneracy-and-numerical}}

\begin{figure}
\centering{}\includegraphics[width=0.9\columnwidth]{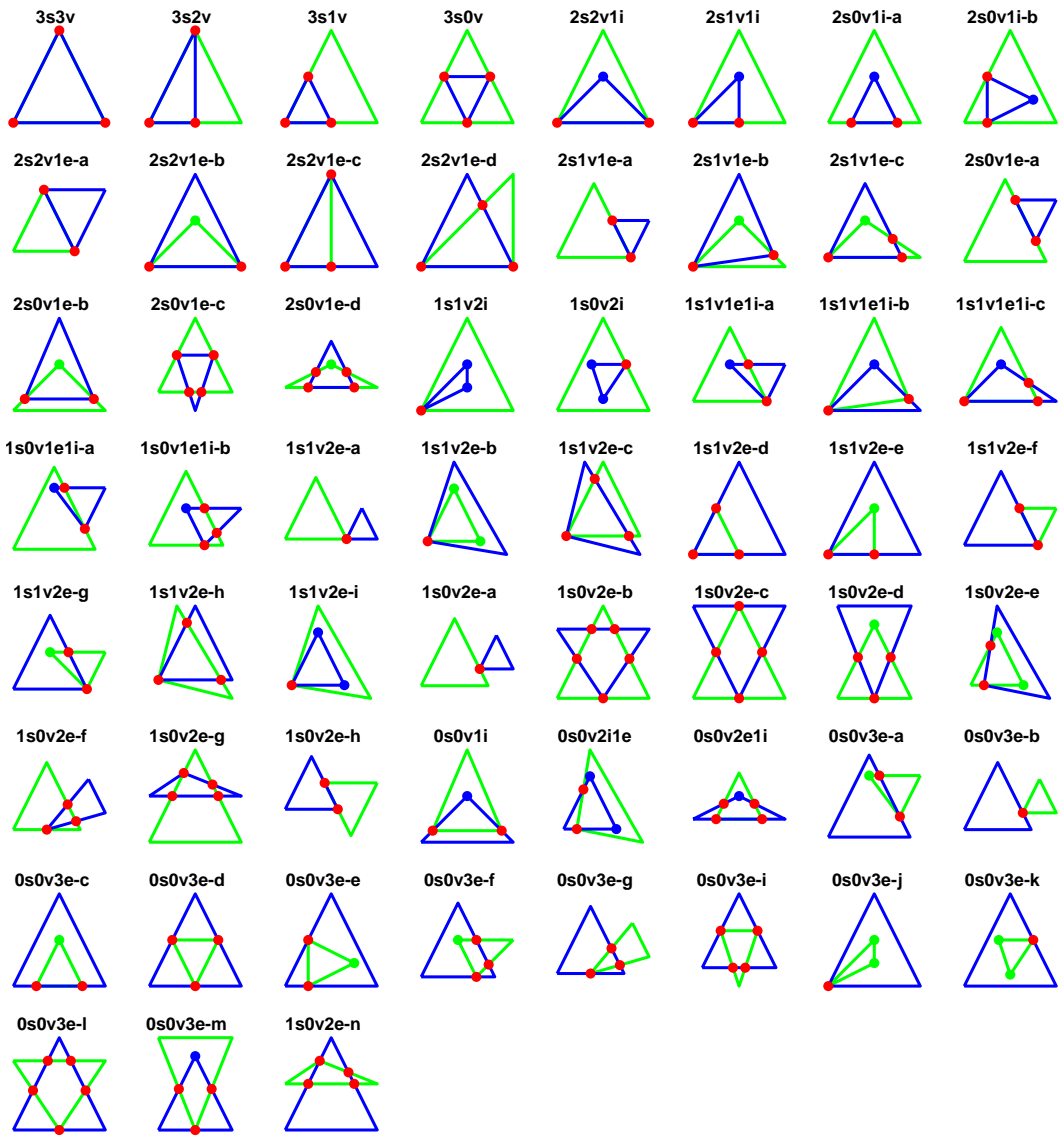}\caption{59 topologically different degenerate cases of triangles with one
or more vertices of $T_{1}$ (blue) falling on sides of $T_{2}$ (green),
or viceversa. A possible classification, hinted in the labels, may
count the number of vertices of $T_{1}$ falling on sides of $T_{2}$
(suffix $s$), the number of common vertices (suffix $v$), the number
of other vertices of $T_{1}$ internal or external to $T_{2}$ (suffixes
$i$ or $e$); but these indicators alone are not exhaustive (hence
the lowercase letters appended to the label).\label{fig:59-onside}}
\end{figure}

Problems arise for triangles which have some colinear side, or simply
vertices of one triangle falling on the sides of the other. Figure
\ref{fig:59-onside} displays 59 topologically different configurations,
and is possibly not even exhaustive of them. Such cases have to be
treated with care in the numerical computation, because their identification
requires the simultaneous satisfaction of more than a single equality
condition. For instance, if one vertex of $T_{1}$ falls onto a side
of $T_{2}$, three conditions are to be true: the vertex must belong
to the side of the second triangle, and two sides of the first triangle
must intersect that side, both exactly in that point. For our purposes,
one of the triangles will be a half pixel in the destination image,
and the other a warped half pixel of the original image. For a generic
functional mapping between $(x,y)$ and $(X,Y)$, expressed by algebraic
or transcendental functions, such cases may be extremely rare; however,
colinear points will be very frequent for transformations like grid
sub or oversampling by integer factors, or rotation by notable angles,
which are indeed among the most typical test cases. A reliable algorithm
for warping must be capable to treat them adequately as well.

At numerical precision, due to the propagation of truncation errors,
exact equality conditions may often be violated; moreover, they may
be violated in a way which is topologically inconsistent. For example,
it may result numerically that one vertex of $T_{1}$ falls on a side
of $T_{2}$, whereas the sides of $T_{1}$ originating from that vertex
may not appear to intersect $T_{2}$, or, to intersect it in numerically
different points than the vertex in question. Additionally, extreme
image warpings around singular points of the coordinate transformation
can produce pathological triangles with nearly-colinear vertices.
The numerical identification of intersections between sides can also
produce bogus results for them. 

A way of coping with nearly degenerate cases would be to evaluate
the algebraic conditions which define whether a point is interior
or exterior to a triangle, or whether two segments intersect, within
preassigned tolerances, larger than typical truncation errors \cite{Jules2014,EricsonChrister2005RCD}.
The problem in that is that the resolution of the ambiguity usually
requires a compatible set of decisions for more than a single test.
If for instance one of the barycentric coordinates of a point is found
to be nearly zero, implying that the point lies on the side of the
test triangle, it is not possible to consider systematically that
point as interior or exterior to the triangle, by itself. The decision
depends on which intersections among sides should be counted or not,
in order to reconduce the limit case to one of the 17 basic ones of
figure \ref{fig:The-seventeen-topologically}.

\subsection{Topological approach\label{subsec:Topological-approach}}

If it was not for the possible degeneration, we could compute the
area of $T_{1}\cap T_{2}$ in each of the 17 cases mentioned in section
\ref{subsec:Enumeration-of-possible} identifying in each configuration
the composing trianglets (up to four) which are to be considered.
Such trianglets are highlighted by pink dashed lines in Fig.~\ref{fig:The-seventeen-topologically}.
In this approach, the case is first identified counting the number
of points of $T_{1}$ internal to $T_{2}$, and then determining the
number and the coordinates of the intersections only among the sides
required. In some of the cases, the knowledge of the internal points
makes the computation of some intersections among sides unnecessary,
saving operations. For instance, in case \textbf{2210} it is known
a priori that one side of $T_{1}$ is all contained in $T_{2}$, and
doesn't intersect any of its sides. Then, with conditional code which
treats each case differently, the relevant trianglets are singled
out and their areas added. This may be, for some of the cases, computationally
more economical than with the approach described in the next section.
A sample Matlab implementation of this method is given in the file
\texttt{areaTriangleIntersection.m} included in the Supplementary
Material (directory \texttt{Triangles/topologicalIntersection/}).
However, the the procedure is incomplete if degenerate cases are not
treated, identified and cast into one of the 17 basic patterns. While
this is possible, it requires detailed case-by-case code, branching
through all possibilities (see the function \texttt{intersectTriangles2.m}
in the same directory). That would lead to a code complicate and difficult
to be maintained, outweighing the minimal performance gain which could
result from it.

\subsection{Brute force approach\label{subsec:Brute-force-approach}}

To avoid the classification of all potentially degenerate triangle
overlap cases, we adopt a simpler approach. We start noting that the
vertices of the polygon $T_{1}\cap T_{2}$ are points which always
belong to at least one of three sets: the vertices of $T_{1}$ which
are internal to $T_{2}$, the vertices of $T_{2}$ internal to $T_{1}$,
and the intersections between sides of $T_{1}$ and $T_{2}$. First,
a list of such points (at most $12=3+3+6$) is compiled using barycentric
coordinates for the computations. In the degenerate cases, some of
these points can appear in more than one of the subsets: for example
a point of $T_{1}$ on a side of $T_{2}$ can appear both as an internal
point and as the intersection of two different sides of $T_{1}$ with
$T_{2}$. The list is therefore pruned, eliminating duplicate points
which coincide within a given numerical tolerance. The area of the
polygon is then computed from the pruned list of $N$ vertices $\left\{ \vec{x}_{k}\right\} =\left\{ \left(x_{k},y_{k}\right)\right\} $
according to algorithm \ref{alg:polygon-area}, which is robust to
nearly coincident or colinear vertices, and economical in terms of
operations.

\begin{algorithm}[h]
if $N<3$, $\mathcal{A}\left[\left\{ \right\} \right]=0$

if $N=3$, 
\[
\mathcal{A}\left[\left\{ \vec{x}_{1},\vec{x}_{2},\vec{x}_{3}\right\} \right]=\frac{\left|\left(x_{3}-x_{1}\right)\left(y_{2}-y_{1}\right)-\left(x_{2}-x_{1}\right)\left(y_{3}-y_{1}\right)\right|}{2}
\]

if $N>3$:
\begin{enumerate}
\item the center point $\vec{x}_{M}=\frac{1}{N}\sum_{k=1}^{N}\vec{x}_{k}$
is computed\label{enu:polygon-area:centerpoint}
\item ray angles from the center are computed, $\varphi_{k}=\tan^{-1}\frac{y_{M}-y_{k}}{x_{M}-x_{k}}$\label{enu:polygon-area:ray-angles}
\item the set $\left\{ \vec{x}_{k}\right\} $ is sorted in order of increasing
$\varphi_{k}$ \label{enu:polygon-area:vertex-sort}
\item $\vec{x}_{1}$ is taken as a pivot, and for $3\le k\le N$ the $N-2$
triangle areas 
\[
\mathcal{A}\left[\left\{ \vec{x}_{1},\vec{x}_{k-1},\vec{x}_{k}\right\} \right]=\frac{\left(x_{k}-x_{1}\right)\left(y_{k-1}-y_{1}\right)-\left(x_{k-1}-x_{1}\right)\left(y_{k}-y_{1}\right)}{2}
\]
 are computed and summed.\label{enu:polygon-area:triangles}
\end{enumerate}
\caption{area of the intersection polygon\label{alg:polygon-area}}
\end{algorithm}

Sorting the vertices in cyclic order is required for $N>3$ (steps
\ref{enu:polygon-area:centerpoint}--\ref{enu:polygon-area:vertex-sort}
of algorithm \ref{alg:polygon-area}), since the pruned list of vertices
is not guaranteed to be ordered, by construction. This adds a computational
cost of two divisions by $N$ (step \ref{enu:polygon-area:centerpoint}),
$N$ evaluations of \texttt{atan2()} (step \ref{enu:polygon-area:ray-angles})
and a sort operation of a list of four to six floating numbers (step
\ref{enu:polygon-area:vertex-sort}). Conditional code to list the
internal points and the intersections in a proper order, on the other
hand, would be more convoluted. As an aside, step \ref{enu:polygon-area:triangles}
involves only $2\left(N-2\right)$ multiplications, whereas the standard
shoelace algorithm for computing the area of the polygon would require
$2N$. The cyclic order of the vertices, and the convexity of the
polygon itself, guarantee that $\left(x_{k}-x_{1}\right)\left(y_{k-1}-y_{1}\right)-\left(x_{k-1}-x_{1}\right)\left(y_{k}-y_{1}\right)>0$
for all $k$. There would be other algorithmic possibilities to achieve
cyclic sorting, avoiding the evaluation of \texttt{atan2()}; (compare
for instance the function \texttt{polygonArea2.m} given in Supplementary
Material, which uses it, with \texttt{polygonArea3.m}, which does
not, both in directory \texttt{Triangles/}); however, as for Matlab
is concerned, the first option is faster).

The advantage of this procedure is that a single threshold identification
criterion is applied to the list of points obtained, without the need
of pondering the compatibility between resolutions of internality
and crossing in the limit cases. Conversely, the effect of including
or excluding some nearly coincident polygon vertices, dependent on
the threshold value chosen, amounts only to adding or not some nearly
null area contributions.

This algorithm is implemented in the file \texttt{areaTriangleIntersection2.m}
included in the Supplementary Material (directory \texttt{Triangles/}),
and has been used to process the images shown in this paper.

\section{Computational performance\label{app:Computational-performance}}

The algorithms described in this paper have been implemented in Matlab,
and the code is provided as Supplementary Material of this paper.
Coding has paid some amount of attention to efficiency and good programming
practice, but ultimate performance has not been sought for itself.

Some CPU times of the algorithm presented, vs.~the much faster resampling
interpolation of the inverse map, are presented in Table \ref{tab:timings}.
Timings were obtained on a 12 core Intel Xeon® W-2135 CPU with 3.70GHz
clock, using Matlab 2020a. Tests runs on square images defined on
unit square coordinates, similar to that reported in section \ref{subsec:Comparison-with-resampling}
were executed at different resolutions, and the average of 10 area
warp iterations and 1000 interpolation iterations is recorded. The
$\sin$ mapping of Eq.~(\ref{eq:sin-mapping}) was chosen, having
an analytical inverse and being bijective on the whole unit square.

\begin{table}[h]
\noindent \caption{Timings for area warping vs.~resampling.\label{tab:timings}}
\noindent \begin{centering}
\begin{tabular}{|c|c|c|c|c|}
\hline 
$I_{1}$ resolution & $I_{2}$ resolution & area warp, ms & \#$B_{lm,ij}^{a}\neq0$ & bilinear resampling, ms\tabularnewline
\hline 
\hline 
$64\times64$ & $100\times100$ & 251.6 & 26244 & 0.313\tabularnewline
\hline 
$64\times64$ & $1000\times1000$ & 4594 & 1127844 & 10.75\tabularnewline
\hline 
$512\times512$ & $100\times100$ & 3540 & 372100 & 2.43\tabularnewline
\hline 
$512\times512$ & $1000\times1000$ & 10748 & 2280100 & 13.43\tabularnewline
\hline 
\end{tabular}
\par\end{centering}
\end{table}

These timings are to be taken only as somewhat indicative of a general
trend. In our implementation of the area warping, the core routine
for computing $\mathcal{A}\left[T_{lm}\cap f\left(t_{ij}\right)\right]$
using eq.~(\ref{eq:barycrossing}) and Algorithm \ref{alg:polygon-area}
is compiled into a \texttt{mex} file, for higher efficiency, but all
other parts of the code are executed in a loop by the Matlab interpreter.
This includes the index referencing to $t_{ij}$ and $T_{lm}$, as
well as the computation of $f\left(t_{ij}\right)$. On the other hand,
the outer loop on $i,j$ of algorithm \ref{alg:Intensity-transformation}
described in §\ref{sec:Algorithm-layout} can be easily parallelized,
using Matlab's \texttt{parfor} construct. In contrast, when we look
at traditional resampling, we are comparing with a single call of
\texttt{interp2}, which is certainly well optimized internally, and
involves much less operations. It is thus not too constraining, at
this stage, to observe that the area warp procedure is some three
orders of magnitude slower than the usual technique. Moreover, the
computational effort for the area warp is expected to be dependent
on the amount of stretching and destination domain coverage caused
by the particular mapping $f$, which affects the sparsity of the
resulting matrix $B_{lm,ij}^{a}$, and even by the number of sides
of each individual triangle intersection generated.

\section*{Acknowledgments}

The author wishes to thank Eran Ofek for fruitful discussion on the
content of this paper.


\begin{thebibliography}{10}

\bibitem{Amanatiadis2009}
{\sc A.~Amanatiadis and I.~Andreadis}, {\em A survey on evaluation methods for
  image interpolation}, Measurement Science and Technology, 20 (2009),
  p.~104015, \url{https://doi.org/10.1088/0957-0233/20/10/104015}.

\bibitem{Beg2005}
{\sc M.~F. Beg, M.~I. Miller, A.~Trouv\'e, and L.~Younes}, {\em Computing large
  deformation metric mappings via geodesic flows of diffeomorphisms},
  International Journal of Computer Vision, 61 (2005), pp.~139--157,
  \url{https://doi.org/10.1023/B:VISI.0000043755.93987.aa}.

\bibitem{Beier1992}
{\sc T.~Beier and S.~Neely}, {\em Feature-based image metamorphosis}, SIGGRAPH
  Comput. Graph., 26 (1992), pp.~35--42,
  \url{https://doi.org/10.1145/142920.134003}.

\bibitem{opencv_library}
{\sc G.~Bradski}, {\em {The OpenCV Library}}, Dr. Dobb's Journal of Software
  Tools,  (2000).

\bibitem{Chiang1998}
{\sc M.-C. Chiang}, {\em Imaging-consistent warping and super-resolution}, PhD
  thesis, Columbia University, 1998,
  \url{https://www.proquest.com/docview/304435854}.

\bibitem{Coxeter1969}
{\sc H.~S.~M. Coxeter}, {\em Introduction to Geometry}, Wiley, 2nd ed.~ed.,
  1969.

\bibitem{Elsheikh2014}
{\sc A.~H. Elsheikh and M.~Elsheikh}, {\em A reliable triangular mesh
  intersection algorithm and its application in geological modelling},
  Engineering with Computers, 30 (2014), pp.~143--157,
  \url{https://doi.org/10.1007/s00366-012-0297-3}.

\bibitem{EricsonChrister2005RCD}
{\sc C.~Ericson}, {\em Real-Time Collision Detection}, The Morgan Kaufmann
  series in interactive 3D technology, CRC Press, London, 2005,
  \url{https://learning.oreilly.com/library/view/real-time-collision-detection/9781558607323/}.

\bibitem{Fant1986}
{\sc K.~M. Fant}, {\em A nonaliasing, real-time spatial transform technique},
  IEEE Computer Graphics and Applications, 6 (1986), pp.~71--80,
  \url{https://doi.org/10.1109/mcg.1986.276613}.

\bibitem{cgal:fwzh-rbso2-20b}
{\sc E.~Fogel, O.~Setter, R.~Wein, G.~Zucker, B.~Zukerman, and D.~Halperin},
  {\em {2D} regularized boolean set-operations}, in {CGAL} User and Reference
  Manual, {CGAL Editorial Board}, {5.3}~ed., 2021,
  \url{https://doc.cgal.org/5.3/Manual/packages.html#PkgBooleanSetOperations2}.

\bibitem{Fruchter2002}
{\sc A.~S. Fruchter and R.~N. Hook}, {\em Drizzle: A method for the linear
  reconstruction of undersampled images}, Publications of the Astronomical
  Society of the Pacific, 114 (2002), pp.~144--152,
  \url{https://doi.org/10.1086/338393}.

\bibitem{Getreuer2011}
{\sc P.~Getreuer}, {\em {Linear Methods for Image Interpolation}}, {Image
  Processing On Line}, 1 (2011), pp.~238--259,
  \url{https://doi.org/10.5201/ipol.2011.g_lmii}.

\bibitem{GHOSH20161}
{\sc D.~Ghosh and N.~Kaabouch}, {\em A survey on image mosaicing techniques},
  Journal of Visual Communication and Image Representation, 34 (2016),
  pp.~1--11, \url{https://doi.org/10.1016/j.jvcir.2015.10.014}.

\bibitem{Glasbey1998}
{\sc C.~A. Glasbey and K.~V. Mardia}, {\em A review of image-warping methods},
  Journal of Applied Statistics, 25 (1998), pp.~155--171,
  \url{https://doi.org/10.1080/02664769823151}.

\bibitem{Guigue2003}
{\sc P.~Guigue and O.~Devillers}, {\em Fast and robust triangle-triangle
  overlap test using orientation predicates}, Journal of Graphics Tools, 8
  (2003), pp.~25--32, \url{https://doi.org/10.1080/10867651.2003.10487580}.

\bibitem{Han2005}
{\sc D.~Han}, {\em Real-time digital image warping for display distortion
  correction}, in Image Analysis and Recognition, M.~Kamel and A.~Campilho,
  eds., Berlin, Heidelberg, 2005, Springer Berlin Heidelberg, pp.~1258--1265,
  \url{https://doi.org/10.1007/11559573_152}.

\bibitem{Heckbert1989}
{\sc P.~S. Heckbert}, {\em Fundamentals of texture mapping and image warping},
  master's thesis, Dept. of Electrical Engineering and Computer Science,
  University of California, Berkeley, 1989,
  \url{http://www.cs.cmu.edu/~ph/texfund/texfund.pdf}.

\bibitem{imagemagick}
{\sc {ImageMagick Development Team}}, {\em {ImageMagick}}, 2021,
  \url{https://imagemagick.org} (accessed 2021-01-04).
\newblock Version 7.0.10.

\bibitem{Jules2014}
{\sc C.~Jules}, {\em Accurate point in triangle test}, 2014,
  \url{http://totologic.blogspot.com/2014/01/accurate-point-in-triangle-test.html}.

\bibitem{Lee1998}
{\sc S.~Lee, G.~Wolberg, and S.~Y. Shin}, {\em Polymorph: morphing among
  multiple images}, IEEE Computer Graphics and Applications, 18 (1998),
  pp.~58--71, \url{https://doi.org/10.1109/38.637304}.

\bibitem{Middleton2005}
{\sc L.~Middleton and J.~Sivaswamy}, {\em Hexagonal Image Processing},
  Springer-Verlag, 2005, \url{https://doi.org/10.1007/1-84628-203-9}.

\bibitem{Milanfar2011}
{\sc P.~Milanfar}, ed., {\em Super-Resolution Imaging}, CRC Press, first~ed.,
  2011, \url{https://doi.org/10.1201/9781439819319}.

\bibitem{Modersitzki2004}
{\sc J.~Modersitzki}, {\em Numerical methods for image registration}, Numerical
  mathematics and scientific computation, Oxford University Press, Oxford,
  2004.

\bibitem{Modersitzki2009}
{\sc J.~Modersitzki}, {\em FAIR: Flexible Algorithms for Image Registration},
  Society for Industrial and Applied Mathematics, 2009,
  \url{https://doi.org/10.1137/1.9780898718843.ch3}.

\bibitem{Moeller2004}
{\sc T.~M\"oller}, {\em A fast triangle-triangle intersection test}, Journal of
  Graphics Tools, 2 (2004), pp.~25--30,
  \url{https://doi.org/10.1080/10867651.1997.10487472},
  \url{http://web.stanford.edu/class/cs277/resources/papers/Moller1997b.pdf}.

\bibitem{Parker1983}
{\sc J.~A. {Parker}, R.~V. {Kenyon}, and D.~E. {Troxel}}, {\em Comparison of
  interpolating methods for image resampling}, IEEE Transactions on Medical
  Imaging, 2 (1983), pp.~31--39,
  \url{https://doi.org/10.1109/TMI.1983.4307610}.

\bibitem{radke_2012}
{\sc R.~J. Radke}, {\em Computer Vision for Visual Effects}, Cambridge
  University Press, 2012, \url{https://doi.org/10.1017/CBO9781139019682}.

\bibitem{Reinelt2003}
{\sc M.~Reinelt}, {\em pamscale, in {Netpbm User Manual}}, 2020,
  \url{http://netpbm.sourceforge.net/doc/pamscale.html} (accessed 2021-10-03).

\bibitem{Sabharwal2015}
{\sc C.~L. Sabharwal and J.~L. Leopold}, {\em A triangle-triangle intersection
  algorithm}, Computer Science \& Information Technology (CS \& IT),  (2015),
  \url{https://doi.org/10.5121/csit.2015.51003},
  \url{https://www.airccj.org/CSCP/vol5/csit54203.pdf}.

\bibitem{Sampoli2004}
{\sc M.~L. Sampoli}, {\em An automatic procedure to compute efficiently the
  intersection of two triangles}, Tech. Report 465, Universit\`a di Siena,
  Dipartimento di Scienze Matematiche ed Informatiche, 2004,
  \url{https://www.researchgate.net/publication/265809721_An_Automatic_Procedure_to_Compute_Efficiently_the_Intersection_of_Two_Triangles}
  (accessed 2022-04-27).

\bibitem{Summers2012}
{\sc J.~Summers}, {\em Image{W}orsener - {P}ixel {M}ixing}, 2012,
  \url{http://entropymine.com/imageworsener/pixelmixing/}.

\bibitem{Takeda2007}
{\sc H.~Takeda, S.~Farsiu, and P.~Milanfar}, {\em Kernel regression for image
  processing and reconstruction}, IEEE Transactions on Image Processing, 16
  (2007), pp.~349--366, \url{https://doi.org/10.1109/TIP.2006.888330}.

\bibitem{Thyssen2012}
{\sc A.~Thyssen}, {\em Image{M}agick v6 examples -- distorting images}, 2012,
  \url{http://www.imagemagick.org/Usage/distorts/#area_resample} (accessed
  2020-9-13).

\bibitem{Tropp2006}
{\sc O.~Tropp, A.~Tal, and I.~Shimshoni}, {\em A fast triangle to triangle
  intersection test for collision detection}, Computer Animation and Virtual
  Worlds, 17 (2006), pp.~527--535, \url{https://doi.org/10.1002/cav.115},
  \url{https://cs.nyu.edu/exact/pap/mesh/fast-tri-tri-intersect2006.pdf}.

\bibitem{scikit-image}
{\sc S.~van~der Walt, J.~L. {S}ch\"onberger, J.~{Nunez-Iglesias},
  F.~{B}oulogne, J.~D. {W}arner, N.~{Y}ager, E.~{G}ouillart, T.~{Y}u, and the
  scikit-image contributors}, {\em scikit-image: image processing in {P}ython},
  PeerJ, 2 (2014), p.~e453, \url{https://doi.org/10.7717/peerj.453}.

\bibitem{Velho2009}
{\sc L.~Velho, A.~Frery, and J.~Gomes}, {\em Warping and Morphing}, Springer
  London, London, 2009, pp.~387--412,
  \url{https://doi.org/10.1007/978-1-84800-193-0_15}.

\bibitem{Weber2018}
{\sc A.~G. Weber}, {\em The {USC-SIPI} image database: Version 6}, 2018,
  \url{http://sipi.usc.edu/database/SIPI_Database.pdf} (accessed 2021-10-03).

\bibitem{Wolberg1990}
{\sc G.~Wolberg}, {\em Digital image warping}, IEEE Computer Society Press, Los
  Alamitos, CA, 1990.

\bibitem{Yadid-Pecht2004}
{\sc O.~Yadid-Pecht and R.~Etienne-Cummings}, {\em CMOS Imagers}, 2004,
  \url{https://doi.org/10.1007/b117398}.

\bibitem{Yamaguchi1985}
{\sc F.~Yamaguchi}, {\em A unified approach to interference problems using a
  triangle processor}, SIGGRAPH Comput. Graph., 19 (1985), pp.~141--149,
  \url{https://doi.org/10.1145/325165.325224}.

\bibitem{Yang2017}
{\sc X.~Yang, R.~Kwitt, M.~Styner, and M.~Niethammer}, {\em Quicksilver: Fast
  predictive image registration --- a deep learning approach}, NeuroImage, 158
  (2017), pp.~378--396, \url{https://doi.org/10.1016/j.neuroimage.2017.07.008}.

\bibitem{Zackay2017a}
{\sc B.~Zackay and E.~O. Ofek}, {\em How to {COAAD} images. {I.} optimal source
  detection and photometry of point sources using ensembles of images}, The
  Astrophysical Journal, 836 (2017), p.~187,
  \url{https://doi.org/10.3847/1538-4357/836/2/187}.

\bibitem{Zhang2020}
{\sc S.~Zhang, P.~X. Liu, M.~Zheng, and W.~Shi}, {\em A diffeomorphic
  unsupervised method for deformable soft tissue image registration}, 120,
  p.~103708, \url{https://doi.org/10.1016/j.compbiomed.2020.103708}.

\bibitem{ZITOVA2003977}
{\sc B.~Zitov\'a and J.~Flusser}, {\em Image registration methods: a survey},
  Image and Vision Computing, 21 (2003), pp.~977--1000,
  \url{https://doi.org/10.1016/S0262-8856(03)00137-9}.

\end{thebibliography}
\end{document}